\documentclass[openany]{article}
\usepackage{amssymb}
\usepackage{amsmath}
\usepackage{booktabs}
\usepackage{caption}  
\usepackage{cite}
\usepackage{graphicx} 
\usepackage{xcolor}
\usepackage{url}
\usepackage{authblk}
\usepackage{hyperref}
\usepackage{todonotes}
\usepackage{placeins}
\usepackage{fancyhdr}
\usepackage{float}

\title{Current Pathology Foundation Models are unrobust to Medical Center Differences}

\author[1]{Edwin D.~de Jong}
\author[2]{Eric Marcus}
\author[2]{Jonas Teuwen}
\affil[1]{\tiny Independent research performed in November 2024, after affiliation with Kaiko and prior to affiliation with Aignostics; updated with additional results in January 2025}
\affil[2]{\tiny The Netherlands Cancer Institute Amsterdam (NKI), Antoni van Leeuwenhoek Hospital (AvL)}

\date{}
\begin{document}
\maketitle

\begin{abstract}
Pathology Foundation Models (FMs) hold great promise for healthcare. Before they can be used in clinical practice, it is essential to ensure they are robust to variations between medical centers. We measure whether pathology FMs focus on biological features like tissue and cancer type, or on the well known confounding medical center signatures introduced by staining procedure and other differences.

We introduce the \textbf{Robustness Index}. This novel robustness metric reflects to what degree biological features dominate confounding features. Ten current publicly available pathology FMs are evaluated. We find that all current pathology foundation models evaluated represent the medical center to a strong degree. Significant differences in the robustness index are observed. Only one model so far has a robustness index greater than one, meaning biological features dominate confounding features, but only slightly.

A quantitative approach to measure the influence of medical center differences on FM-based prediction performance is described. We analyze the impact of unrobustness on classification performance of downstream models, and find that cancer-type classification errors are not random, but specifically attributable to \textit{same-center confounders}: images of other classes from the same medical center. We visualize FM embedding spaces, and find these are more strongly organized by medical centers than by biological factors. As a consequence, the medical center of origin is predicted more accurately than the tissue source and cancer type. 

The robustness index introduced here is provided with the aim of advancing progress towards clinical adoption of robust and reliable pathology FMs.
\end{abstract}

\section{Introduction}
Pathology Foundation Models (FMs) have quickly become the dominant approach in current pathology AI. Following Campanella's groundbreaking work that used weakly-supervised learning to scale up machine learning for pathology \cite{campanella2019clinical} and early papers applying Self-Supervised Learning (SSL) in the domain \cite{koohbanani_aug2020,dehaene_dec2020,hipt}, an impressive and quickly growing series of pathology FMs have become available, with more than ten new such models published last year so far alone: \cite{phikon,phikonv2,hibou,virchow,virchow2,uni,sra,hoptimus0,prov_gigapath,medgemini,pathfoundation,conch, remedis, exa_onepath}. 


Several of these models demonstrate remarkable capabilities and can detect patterns that human pathologists struggle to observe from H\&E slides, such as Microsatellite Instability (MSI) and Immunohistochemistry (IHC) biomarkers such as Ki67 and PD-L1  \cite{faa2024artificial,deepsmile,virchow}.

Pathology foundation models thus hold great potential for healthcare by aiding pathologists, for example in routine or large volume, labor intensive tasks. If this promise is to be realized, it is essential that models can be trusted to provide unbiased estimates of a patient's condition.

A particular obstacle to the adoption of pathology FMs in clinical practice is the sensitivity of machine learning models (ML) to staining variations, caused by differences in the staining procedures used by different labs, the staining fluids, and imaging equipment.
It is well known that these image variations can influence pathology ML models, and reduce their ability to generalize to data from laboratories not seen during training \cite{Tellez_2019}. 

Staining procedures vary per laboratory, and are thereby associated with specific medical centers. This leads to a clear risk of bias \cite{biasedai}: if models are sensitive to the medical center from where images originate, then patients from different medical centers will be evaluated differently by such models, possibly leading to different diagnoses and treatment based on irrelevant technical differences between images. To ensure foundation models can be safely introduced into healthcare practice, evaluating and confirming their robustness to image variations that occur in practice is a necessary step.

To assess whether current pathology foundation models provide an objective assessment of a patient's condition, we analyze to what extent medical centers influence the embedding spaces generated by FMs. Our contributions are as follows:

\begin{itemize}
\item A basic yet effective description for the concept of \textit{robustness} in medical ML is suggested, based on the distinction between biological features and confounding features. Models can vary in their robustness to image variations such as noise, color differences, augmentations, and variations between medical centers. 
\item We introduce a novel robustness metric: the Robustness Index, measuring the degree to which biological features dominate confounding features in the neighborhood structure of the embedding space induced by the foundation model.
\item For the first time, a quantitative approach to measure the influence of medical center differences on FM-based prediction performance is described. The approach directly relates prediction errors to same-center confounders: images from the same center as the predicted sample that have a different class, and thereby contribute to incorrect classification.
\item 10 current pathology foundation models are evaluated on their medical center robustness. It is found that current pathology FMs vary widely in their robustness to medical center variations.
\item We suggest that the value of an FM is determined by the relation between (A) its ability to characterize relevant biological information, enabling high prediction performance of biological information in downstream tasks, and (B) its robustness, reflected in its insensitivity to irrelevant non-biological variation such as staining and medical center differences. 
\item To gain further insight into the mappings learned by pathology FMs, we project embeddings to a 2D space for visualization using t-SNE \cite{tsne}. We find that most foundation models show a clear clustering of medical centers in the embedding space; this shows more clearly than the clustering of biological classes.
\end{itemize}

\section{Related Work}

In previous work \cite{kaiko_amld24} presented at AMLD 2024, we visualized the embedding space of ViT models trained on TCGA using DINO. It was found that a 2D t-SNE projection of the embedding space was clustered by medical center, forming the inspiration for the current work.

In simultaneous work from the TU Berlin BIFOLD group \cite{koemen_batch_effects}, batch effects in pathology foundation models were analyzed and shown. \cite{elphick2024latent} analyzes rotation invariance. 
Tellez \cite{Tellez_2019} quantified the effects of data augmentation and stain color normalization in pathology in the context of CNNs, and \cite{ciompi_stain_norm} evaluated the effect of stain normalization in colorectal tissue classification.

\cite{sikaroudi2022hospital} discusses the issue that hospitals are represented in WSI data, and looks into domain generalization for hospital-agnostic image representation learning; \cite{ren2018adversarial} also uses domain adaptation to reduce the influence of staining differences.
\cite{biased_data_ai} draws attention to the existence of medical center signatures in TCGA data specifically, and finds that deep neural networks can predict the acquisition site.

A main consequence of unrobustness is that predictive models based on unrobust representations can lead to biased predictions. \cite{howard2021impact} studies the impact of medical center signatures on deep learning model accuracy and bias. \cite{exa_onepath} notes that embeddings from SSL models tend to cluster by individual WSIs and proposes the EXAONEPath pathology FM to address this, which is included in our evaluation here.

While finalizing this paper, a new relevant article discussing the measurement and optimization of robustness in pathology became available \cite{owkin25_robustness}.

\section{Robustness for Medical Foundation Models}

To clarify what is meant by robustness in this work, we distinguish between \textit{biological} features and \textit{confounding} features. Biological features include any relevant features that reflect the true condition of the patient; the aim and promise of foundation models is to capture these. Confounding features are any irrelevant variations in the input that are not related to true biological differences between samples, but are rather caused by external influences such as staining differences, differences in image capture equipment, image processing pipelines, and noise. Given these notions, we can define robustness as insensitivity to confounding features.

\subsection{Robustness Index}
To gain insight into what a foundation model has learned, we can analyze the embedding space by considering the neighborhood around each embedding, i.e.~the closest embeddings. We consider:
\begin{itemize}
\item How many of the \textit{k} nearest neighbors represent the same biological class, e.g.~tissue type or cancer type, in total across all samples
\item How many of the \textit{k} nearest neighbors represent the same medical center, in total across all samples
\end{itemize}
We define the \textbf{medical center robustness index} as the ratio between these quantities. For other biological classes (e.g.~other diseases, or pharmacogenomic groups) or confounding factors (e.g.~scanner type), robustness indices can be defined analogously.\\

Formally: we define the robustness index $R_k$ for a given dataset $D$ containing $n$ samples as:

\begin{equation}
R_k = \frac{\sum_{i=1}^n \sum_{j=1}^k \mathbf{1}(y_j = y_i)}{\sum_{i=1}^n \sum_{j=1}^k \mathbf{1}(c_j = c_i)}
\end{equation}

Where:

\begin{itemize}
    \item $k$ is the number of nearest neighbors considered; in this work, $k=50$ 
    \item $y_j$ is the biological class of the $j$-th nearest neighbor; $y_i$ is the biological class of the sample $i$
    \item $c_j$ is the medical center of the $j$-th nearest neighbor; $c_i$ is the medical center of the sample $i$
    \item $\mathbf{1}(\cdot)$ is the indicator function:  1 if the condition is true, 0 otherwise
\end{itemize}

The numerator represents the total number of nearest neighbors that have the same biological class across all samples, while the denominator represents the total number of nearest neighbors that are from the same medical center across all samples. Cosine distance is used as the distance metric, as cosine similarity is a common way to evaluate embedding similarity.

As an example, consider an embedding spaced dominated by a confounder, say center, and only minutely influenced by the cancer type. Then, the denominator will be large, since most samples will be surrounded by other samples of the same center; the numerator is small since the embedding space is mostly organized by the confounder. \\ A robustness index of 1 means that biological and confounding information are represented equally strongly. In an idealized scenario, the embedding space is blind to the confounding information and completely organized by the biological signal, yielding $R_k = R_{max}$, where $R_{max}$ is determined by the random chance level for encountering a neighbor from the same medical center or confounding class. In practice, it may not be feasible to completely remove  center information, but one would like to see $R_k >> 1$. For the dataset used in this paper, the Robustness Index is expected to vary between 1/4 for random cancer type prediction in combination with perfect medical center prediction and 4 for the converse.\\

\section{Experimental Setup}
Apart from robustness, another important measure of the quality of a foundation model is reflected in the prediction performance of the downstream models built upon it. In Section \ref{task}, we therefore define a basic classification task.

\subsection{Classification Task: Tissue of Origin / Cancer Type}
\label{task}
A classification task for the cancer types of five TCGA projects is defined: BReast invasive CArcinoma (BRCA), COlon ADenocarcinoma (COAD), LIver Hepatocellular Carcinoma (LIHC), LUng Squamous cell Carcinoma (LUSC) and STomach ADenocarcinoma (STAD). Note that these cancer types have a one-to-one correspondence with five different tissues of origin (breast, colon, liver, lung, stomach); so this task can equivalently be viewed as a tissue of origin classification task.

These particular five cancer types were selected in combination with five medical centers such that for each cancer type, TCGA WSI data from multiple medical centers is available and vice versa, resulting in the following selection of centers: Asterand, Greater Poland Cancer Center (GPCC), ILSBio, International Genomics Consortium (IGC), MSKCC; see table \ref{tab:tss_project}. 

To build the dataset, for each available combination of center and cancer type, 10 WSIs are selected randomly. From each of the resulting WSIs, 10 informative foreground patches representing regular tissue were selected from a larger randomly generated set based on visual inspection, filtering out anomalies and low-information (white) patches. This resulted in a dataset of 2000 patches in total. We name this dataset {\tt TCGA-2k}. In all experiments, 5-fold cross-validation is used to generate validation predictions for this whole dataset. 

\subsection{Control Classification Task: Medical Center Predictions}
\label{control_task}
To evaluate to what extent FM embeddings encode the medical center from which images originate, prediction of the medical center of the image is evaluated as a control classification task.

\begin{table}[H]
    \centering
    \caption{Composition of the {
    \tt TCGA-2k} dataset: Tissue Source Site (TSS), Short Name and Project Code Combinations}
    \begin{tabular}{llllll}
        \toprule
        TSS Short Name & BRCA & COAD & LIHC & LUSC & STAD \\
        \midrule
        Asterand & \textcolor{green}\checkmark & \textcolor{green}\checkmark & \textcolor{green}\checkmark & \textcolor{green}\checkmark & \textcolor{green}\checkmark \\
        GPCC & \textcolor{green}\checkmark & \textcolor{green}\checkmark &  &  & \textcolor{green}\checkmark \\
        IGC & \textcolor{green}\checkmark & \textcolor{green}\checkmark & \textcolor{green}\checkmark & \textcolor{green}\checkmark & \textcolor{green}\checkmark \\
        ILSBio & \textcolor{green}\checkmark & \textcolor{green}\checkmark & \textcolor{green}\checkmark &  & \textcolor{green}\checkmark \\
        MSKCC & \textcolor{green}\checkmark & \textcolor{green}\checkmark &  & \textcolor{green}\checkmark &  \\
        \bottomrule
    \end{tabular}
    \label{tab:tss_project}
\end{table}

\subsection{Downstream Task Learning Algorithm and Setup}

To ensure we evaluate the quality of FM embeddings, rather than the performance learned by a complex downstream model, we use one of the simplest possible downstream model architectures: k-nearest neighbor (\textit{knn}, with \textit{k}=3) unless otherwise specified, using cosine similarity as the distance function. 

For image pre-processing and obtaining embeddings from the model output, the default choices for each model are followed. For further details, see the Appendix \ref{appendix}.

\section{Results}
Ten current pathology models were selected; see Appendix \ref{app:model_selection} for details on the selection. For each model, embeddings were generated for all patches in the dataset. The first result subsection below describes the prediction performance of cancer type and medical center, and a quantitative evaluation of the influence of medical center differences on FM-based predictions.

\subsection{Embedding Space Structure and Robustness Index}

As noted above, for each sample, we can measure the number of neighbors that have the same biological or confounding class as the sample, i.e., whether the neighbor has: 
 \begin{itemize}
     \item the same cancer type as the sample, or
     \item the same medical center
 \end{itemize}

 The same information can be plotted as a function of the neighbor index \textit{k}, by calculating the fraction of samples that have the same biological or confounding class. This way, we can summarize and visualize the neighborhood structure of the entire embedding space. For the knn distance metric, cosine distance is used, as cosine similarity is a common way to evaluate embedding similarity.

\renewcommand{\tablename}{Figure}
\addtocounter{figure}{1}
\addtocounter{table}{-1}
\renewcommand{\thetable}{\arabic{figure}}
\begin{table}[H]
\centering
\begin{tabular}{|c|c|}
\hline
\begin{minipage}[t]{0.45\textwidth}
    \centering
    \includegraphics[width=\linewidth]{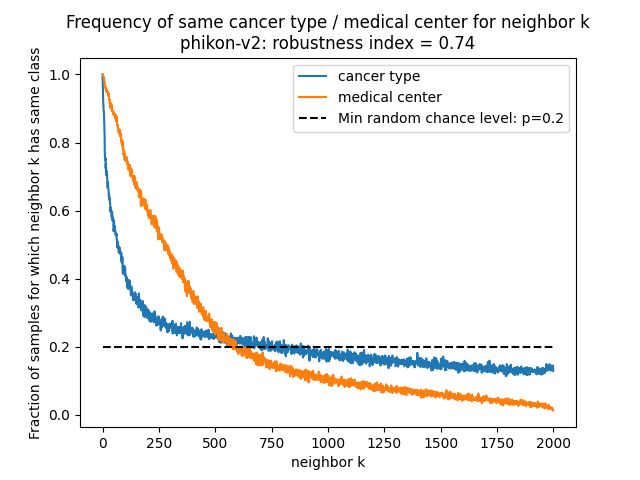}
    \label{fig:freq-same-class-phikon-v2}
\end{minipage}
&
\begin{minipage}[t]{0.45\textwidth}
    \centering
    \includegraphics[width=\linewidth]{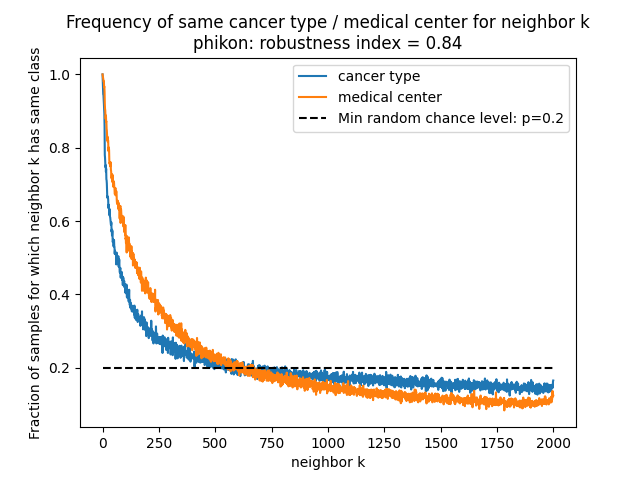}
    \label{fig:freq-same-class-phikon}
\end{minipage}
\\[-8pt]
\hline
\begin{minipage}[t]{0.45\textwidth}
    \centering
    \includegraphics[width=\linewidth]{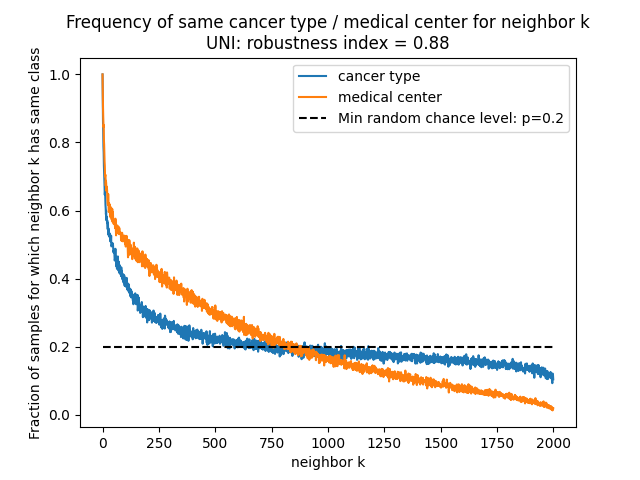}
    \label{fig:freq-same-class-UNI}
\end{minipage}
&
\begin{minipage}[t]{0.45\textwidth}
    \centering
    \includegraphics[width=\linewidth]{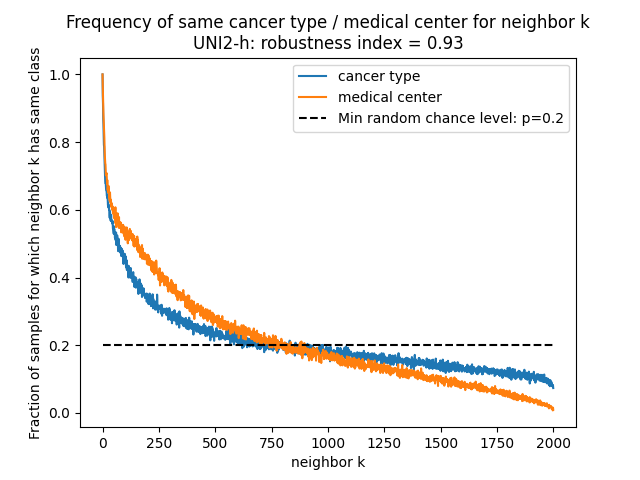}
    \label{fig:freq-same-class-UNI2-h}
\end{minipage}
\\[-8pt]
\hline
\begin{minipage}[t]{0.45\textwidth}
    \centering
    \includegraphics[width=\linewidth]{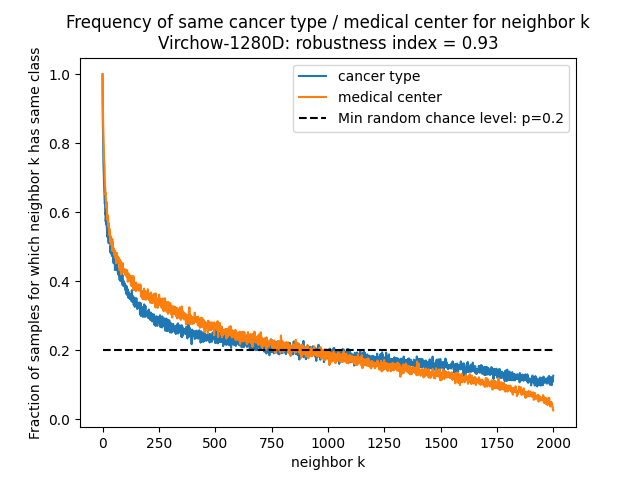}
    \label{fig:freq-same-class-Virchow-1280D}
\end{minipage}
&
\begin{minipage}[t]{0.45\textwidth}
    \centering
    \includegraphics[width=\linewidth]{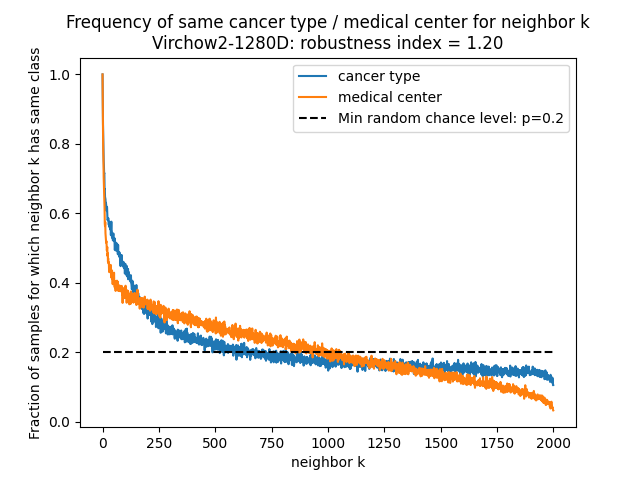}
    \label{fig:freq-same-class-Virchow2-1280D}
\end{minipage}
\\[-8pt]
\hline
\end{tabular}
\caption{Fraction of samples for which the k-th neighbor has the same cancer type (blue) or medical center (orange), in order of increasing robustness. See Appendix \ref{app:fraction-same-class} for all results. For all models, closeness in embedding space is strongly determined by whether the image comes from the same medical center. For all models except Virchow2, the medical center more strongly determines embedding proximity than the cancer type for the nearest 200 neighbors.}
\label{fig:fraction-same-class}
\end{table}
\renewcommand{\tablename}{Table}
\renewcommand{\thetable}{\arabic{table}}

The robustness index can be calculated from these graphs by taking the leftmost values up to $k=50$, taking the averages of the orange and blue lines, and then taking the ratio between these two average values. The robustness index reflects the extent to which biological factors such as cancer type dominate confounding factors such as medical center in the organization of the embedding space.

\begin{figure}
    \centering
    \includegraphics[width=0.7\linewidth]{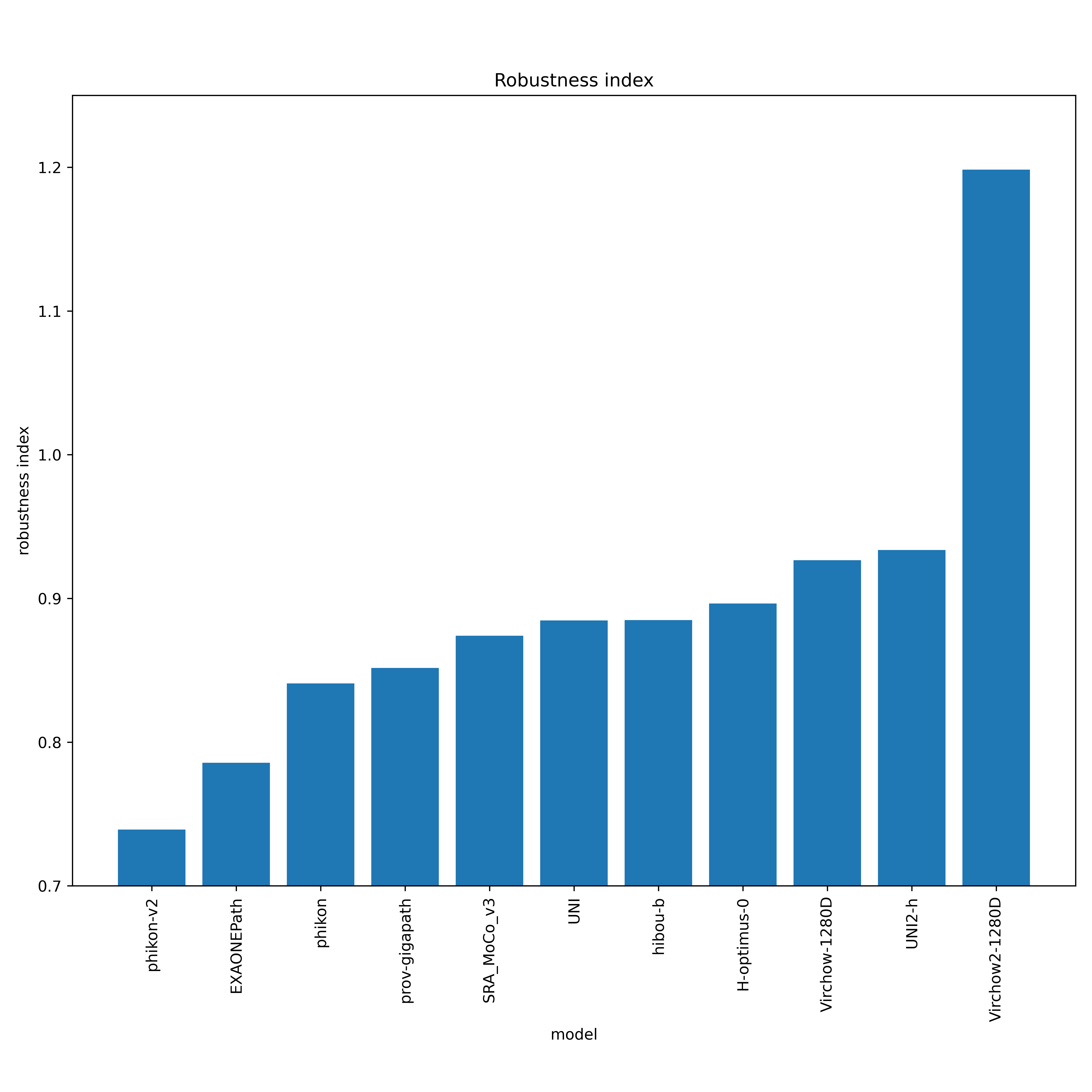}
    \caption{Robustness index for the models evaluated here. }
    \label{fig:robustness-index}
\end{figure}

Figure \ref{fig:fraction-same-class} shows the results, and Figure \ref{fig:robustness-index} summarizes the resulting values of this metric for the models evaluated here. Some observations:
\begin{itemize}
\item According to this analysis, Phikon-v2, an expectedly improved version of Phikon, is less robust than Phikon (0.74 vs 0.84)
\item Uni2-h is more robust than Uni (0.93 vs 0.88)
\item Virchow2 is more robust than Virchow (1.2 vs 0.93), and than all other models. In fact, it is the only model with a robustness index above one, indicating that cancer type information dominates medical center information for the first 50 neighbors. Accordingly, this is the only model for which the blue line is above the orange line for the first 100+ neighbors
\item One might expect the orange and blue lines to level off to an average value around the random chance level above a certain distance. Interestingly, this is not the case; even for the very furthest embeddings, an increase in distance still corresponds to lower probabilities of encountering the same cancer type or medical center. This shows that the organization by cancer type and medical center extends across the whole embedding space, and is a global phenomenon.
\end{itemize}
\subsection{Quantification of the Influence of Medical Center Differences on FM-based Prediction Performance}
\label{fraction_same_center_confounders}
Knn prediction performance was evaluated as follows:
\begin{itemize}
\item For all possible values of \textit{k}, the accuracy of 5-class tissue type / cancer type classification was evaluated using 5-fold cross-validation (green lines).
\item The accuracy of the 5-class medical center classification from which the patch originated was also evaluated (blue lines).
\end{itemize}

In addition to the above common metrics, we aim to measure the influence of the medical center on cancer type classification. To do so, we consider all samples (patches) for which the predicted cancer type class is incorrect.  Given that knn operates by taking the class most common among the sample's \textit{k} nearest neighbors in the training set (as determined by cross-validation here), we can identify the exact set of neighbors that contributed to the incorrect class prediction; this set consists of all neighbors that have the predicted (incorrect) class.

If the FM were completely insensitive to differences between medical centers, the samples that contributed to the incorrect class prediction would be distributed randomly over the centers for that cancer type. To ensure this is the case, we restrict this analysis of the same-center confounders to the two classes that each have data for 5 centers (BRCA and COAD), so that the number of centers for this binary cancer type prediction is equal (5) for all sample points, resulting in a random chance level of occurring for each center of 1/5. Thus, the frequency of any center among these patches is expected to be 1/5.

If, on the other hand, the FM is sensitive to center differences, and tends to organize its embedding space by clustering patches from the same medical center together, then the set of neighboring patches described above will be more likely to come from the same center as the predicted sample. If the fraction of neighbors with the incorrect class that have the same center as the sample is higher than chance level (1/5), this indicates that the FM is sensitive to medical center differences, and that this sensitivity contributes to misclassification. We name such patches \textit{same-center confounders}, as they confound the class prediction.

The figures in Figure \ref{fig:fraction-confounders-sel} and Figure \ref{fig:fraction-confounders-together} show the results; see Appendix Section \ref{fig:fraction-confounders-full} for the complete set of these graphs. 
All models show clear sensitivity to the medical center; the incorrectly classified nearest neighbors are more likely to come from the same center as the predicted samples, in some cases to an extreme extent. E.g.~for the Phikon-v2 model (blue line at the top in Fig.~\ref{fig:fraction-confounders-together}), the closest neighbors of the incorrectly predicted class are from the same center in more than 95\% of the cases. In other words, the embedding space is structured such that a knn classifier will effectively base its classification on whether the patch is from the same center, rather than whether it's from the same biological class.
\renewcommand{\tablename}{Figure}
\addtocounter{figure}{1}
\addtocounter{table}{-1}
\renewcommand{\thetable}{\arabic{figure}}
\begin{table}[H]
\centering
\begin{tabular}{|c|c|}
\hline
\begin{minipage}[t]{0.4\textwidth}
    \centering
    \includegraphics[width=\linewidth]{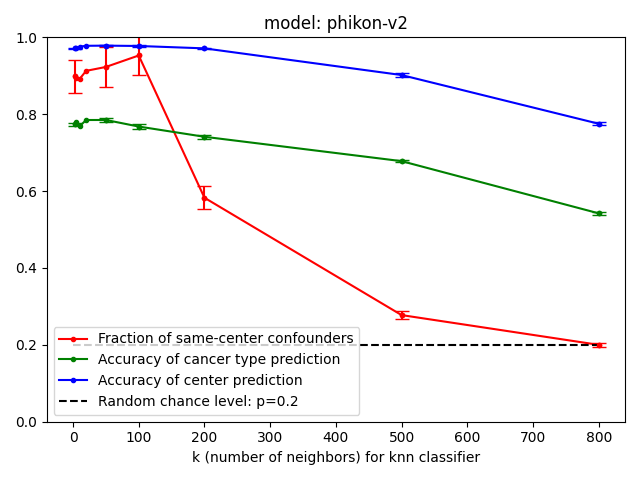}
    \label{fig:fraction-confounders-phikon-v2}
\end{minipage}
&
\begin{minipage}[t]{0.4\textwidth}
    \centering
    \includegraphics[width=\linewidth]{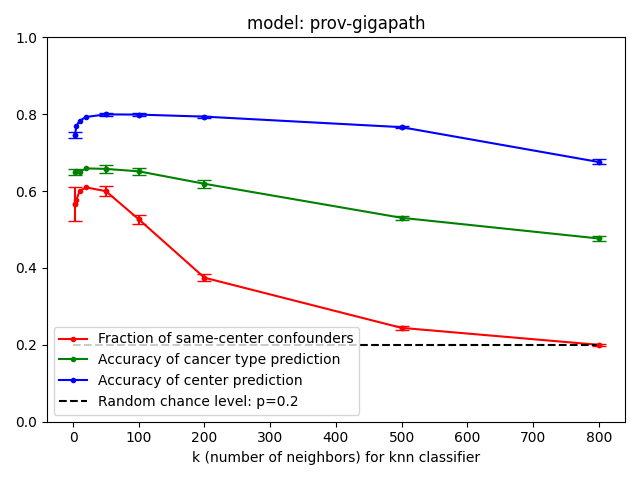}
    \label{fig:fraction-confounders-prov-gigapath}
\end{minipage}
\\[-8pt]
\hline
\begin{minipage}[t]{0.4\textwidth}
    \centering
    \includegraphics[width=\linewidth]{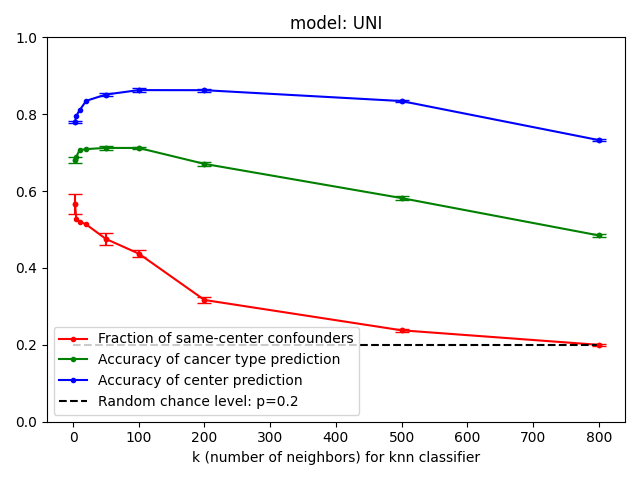}
    \label{fig:fraction-confounders-UNI}
\end{minipage}
&
\begin{minipage}[t]{0.4\textwidth}
    \centering
    \includegraphics[width=\linewidth]{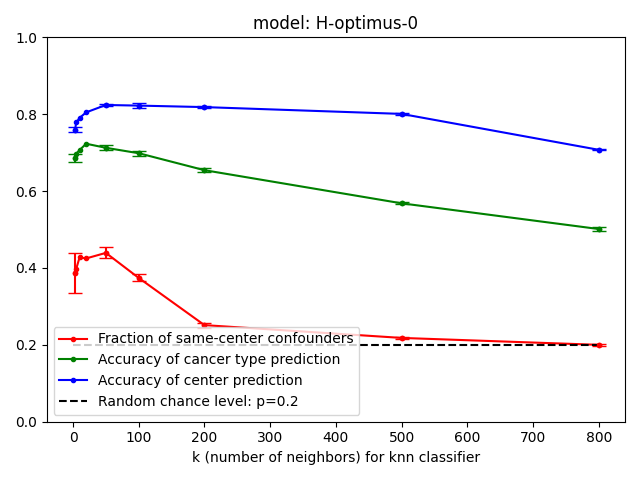}
    \label{fig:fraction-confounders-H-optimus-0}
\end{minipage}
\\[-8pt]
\hline
\begin{minipage}[t]{0.4\textwidth}
    \centering
    \includegraphics[width=\linewidth]{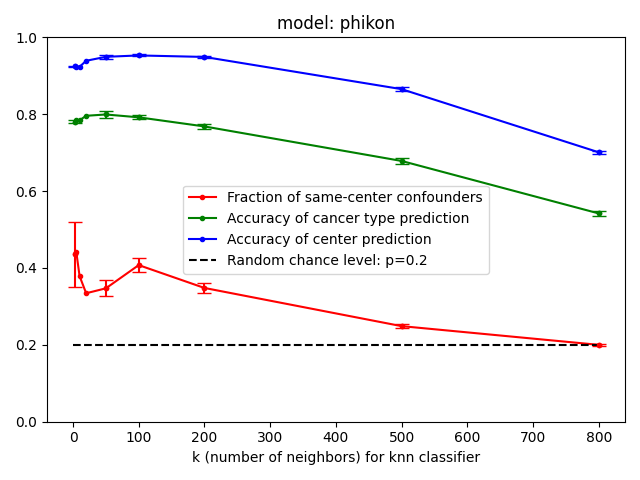}
    \label{fig:fraction-confounders-phikon}
\end{minipage}
&
\begin{minipage}[t]{0.4\textwidth}
    \centering
    \includegraphics[width=\linewidth]{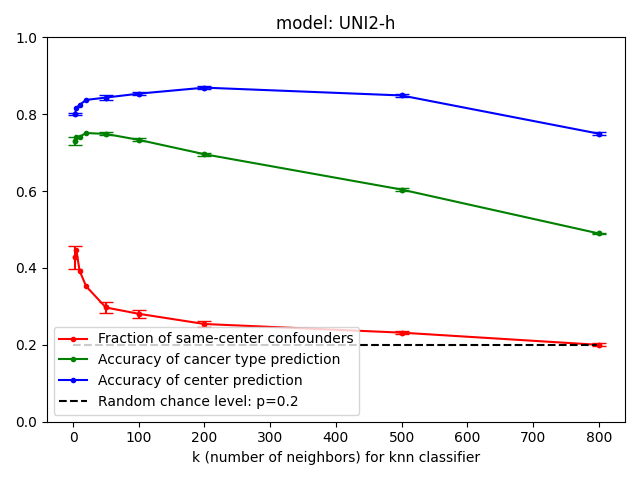}
    \label{fig:fraction-confounders-UNI2-h}
\end{minipage}
\\[-8pt]
\hline
\begin{minipage}[t]{0.4\textwidth}
    \centering
    \includegraphics[width=\linewidth]{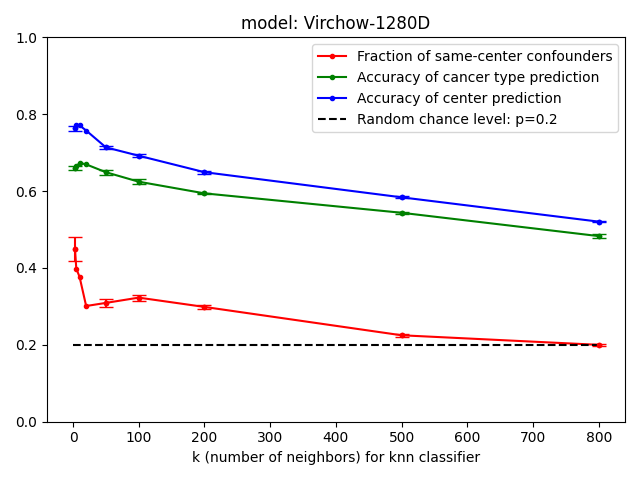}
    \label{fig:fraction-confounders-Virchow-1280D}
\end{minipage}
&
\begin{minipage}[t]{0.4\textwidth}
    \centering
    \includegraphics[width=\linewidth]{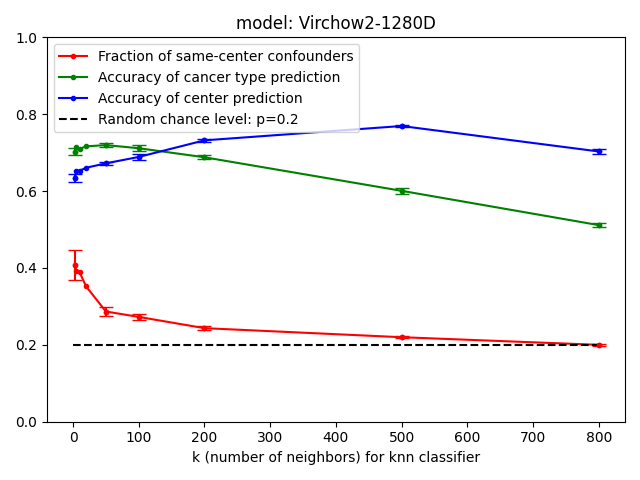}
    \label{fig:fraction-confounders-Virchow2-1280D}
\end{minipage}
\\[-8pt]
\hline
\end{tabular}
\caption{Fraction of same-center confounders for neighbors from the incorrectly predicted class (red);  accuracy of tissue / cancer type prediction (green);  accuracy of medical center prediction (blue), sorted by order of increasing center-robustness for selected FMs. All models show a substantial and significant influence of same-center confounders. See Appendix \ref{app:fraction-confounders-full} for a complete overview.  }
\label{fig:fraction-confounders-sel}
\end{table}
\renewcommand{\tablename}{Table}
\renewcommand{\thetable}{\arabic{table}}


\begin{figure}
    \centering
    \includegraphics[width=0.5\linewidth]{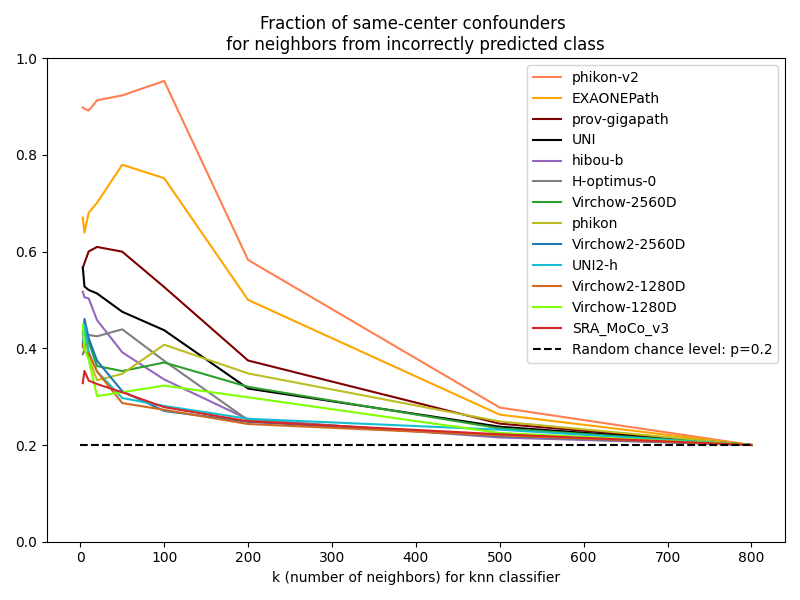}
    \caption{Fractions of same-center confounders for all models. All models are sensitive to these differences, some to a very high degree.}
    \label{fig:fraction-confounders-together}
\end{figure}

\subsection{Visualization of the Embedding Space}
To gain insight into the embedding spaces learned by the models, we use t-SNE \cite{tsne} to project the high-dimensional embedding vectors to 2D. This results in 2D plots where each patch is represented by a dot in 2D space. The t-SNE method is run in an unsupervised manner; i.e.~no label information about cancer type or medical center is used to obtain the 2D embeddings. 

Given the 2D patch embeddings, we can color the embeddings using meta-information about the patches. Figures \ref{fig:colored_embeddings_1} and \ref{fig:colored_embeddings_2} show colorings of the 2D embeddings by cancer type (left column) and medical center (right column); note that the patch locations (the locations of the dots) in these left and right columns are identical.

The figures on the left show some degree of clustering by cancer type. No model achieves perfect separation; this may be unattainable, as patches are selected randomly from the foreground, and some patches may not contain sufficient information to identify the tissue of origin or the corresponding cancer type. 

The figures on the right colored by medical center in general show increased clustering. The coloring for phikon-v2 shows extreme, almost perfect clustering by medical center; the medical center can be predicted with near-perfect accuracy based on the 2D embedding space location alone. This explains the high sensitivity to medical centers seen in the above result section \ref{fraction_same_center_confounders}.

\renewcommand{\tablename}{Figure}
\addtocounter{figure}{1}
\addtocounter{table}{-1}
\renewcommand{\thetable}{\arabic{figure}}
\begin{table}[H]
\centering
\begin{tabular}{|l|c|c|}
\hline
\textbf{Model} & \textbf{Cancer Type} & \textbf{Medical Center} \\
\hline
hibou-b &
\includegraphics[width=0.34\linewidth]{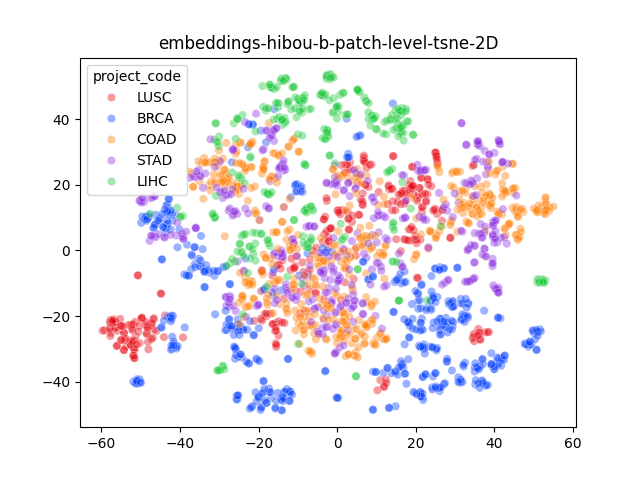} &
\includegraphics[width=0.34\linewidth]{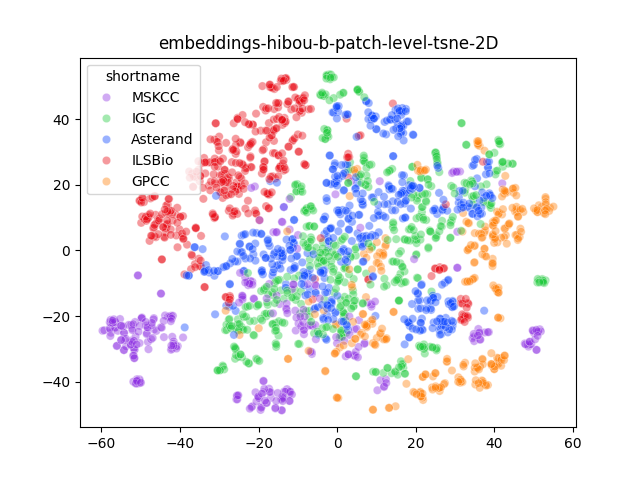} \\
\hline
phikon &
\includegraphics[width=0.32\linewidth]{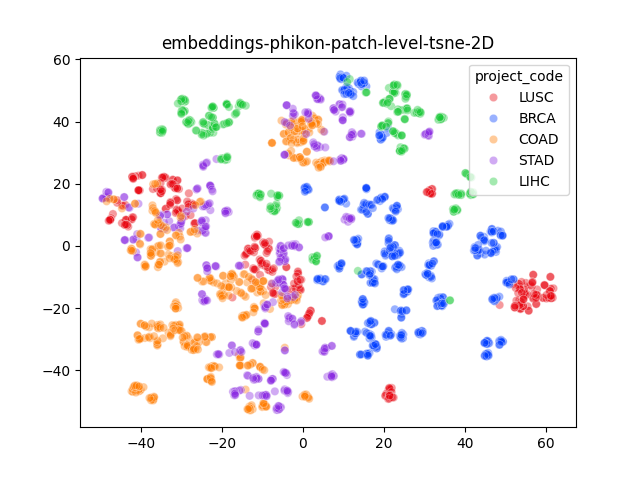} &
\includegraphics[width=0.32\linewidth]{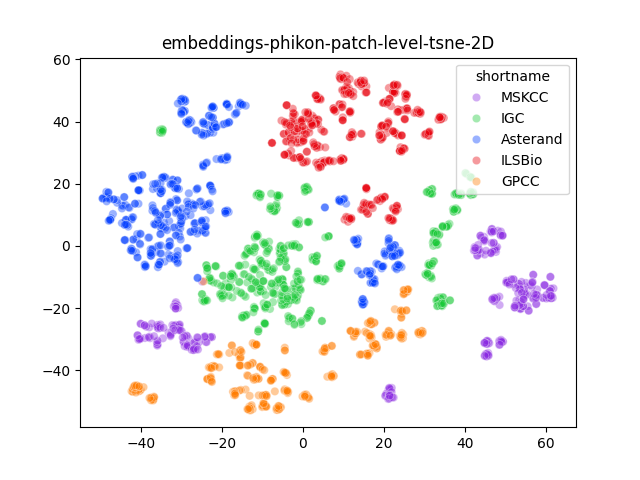} \\
\hline
phikon-v2 &
\includegraphics[width=0.32\linewidth]{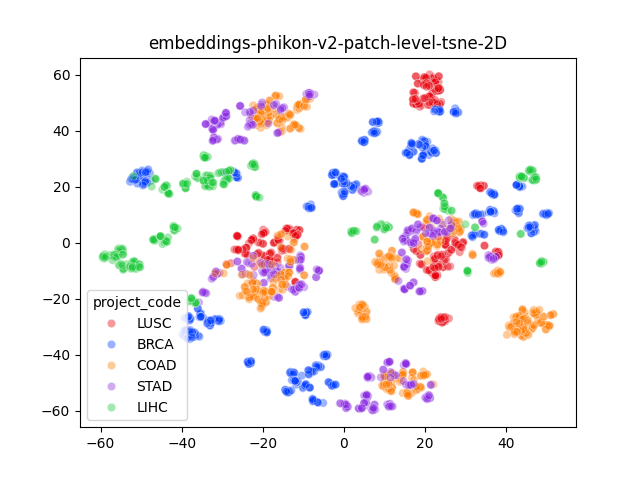} &
\includegraphics[width=0.32\linewidth]{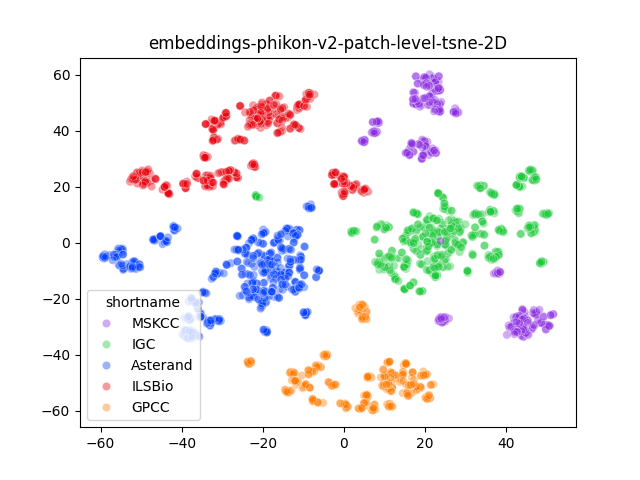} \\
\hline
EXAONEPath &
\includegraphics[width=0.32\linewidth]{
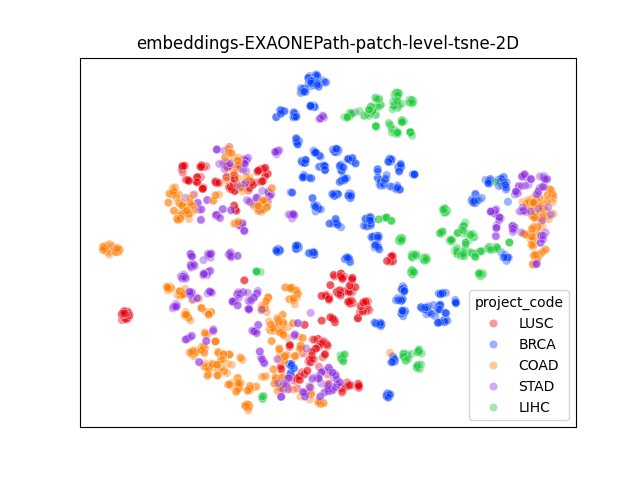
} &
\includegraphics[width=0.32\linewidth]{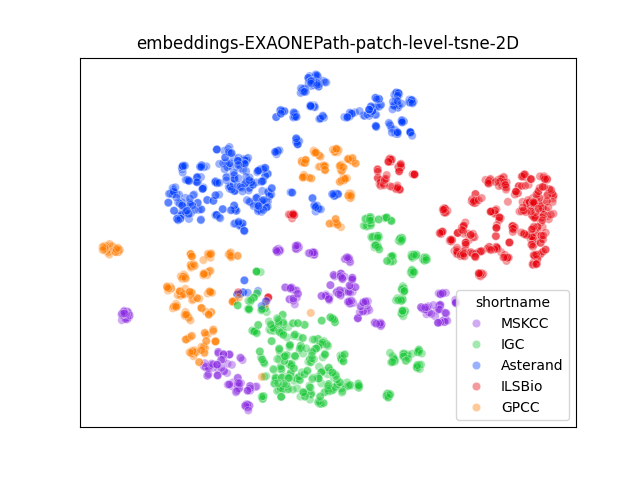} \\
\hline
UNI &
\includegraphics[width=0.32\linewidth]{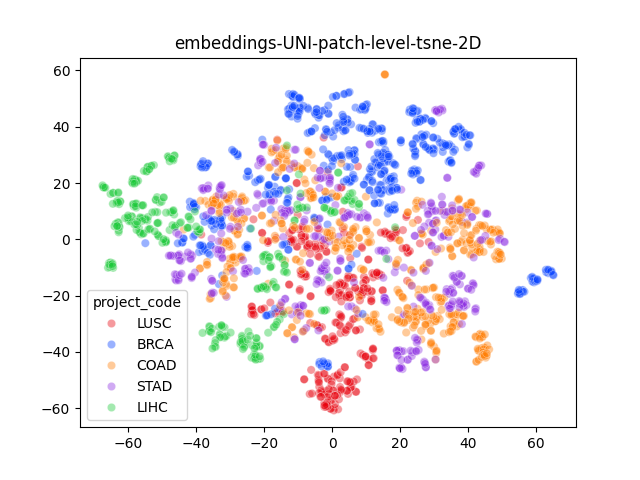} &
\includegraphics[width=0.32\linewidth]{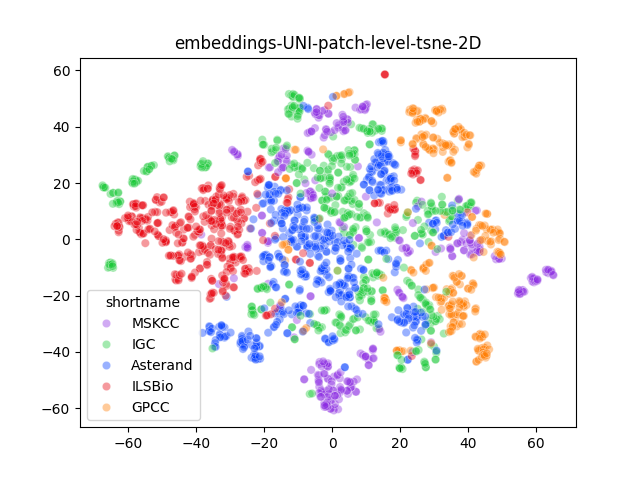} \\
\hline
UNI2-h &
\includegraphics[width=0.32\linewidth]{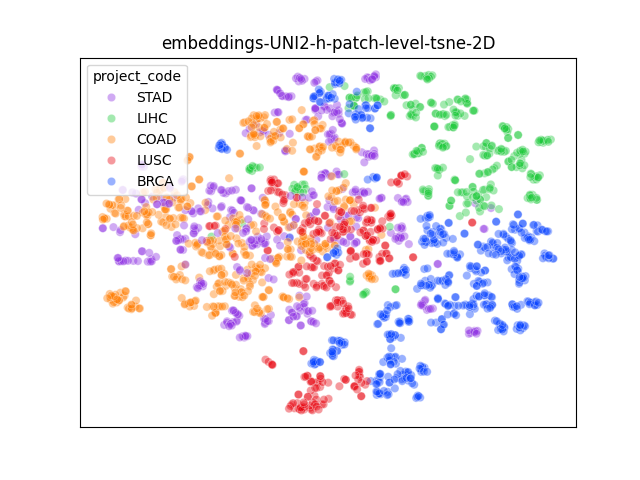} &
\includegraphics[width=0.32\linewidth]{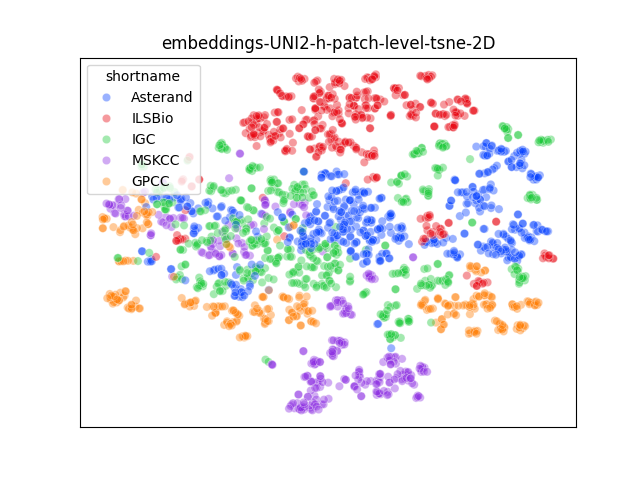} \\
\hline
SRA\_MoCo\_v3 &
\includegraphics[width=0.32\linewidth]{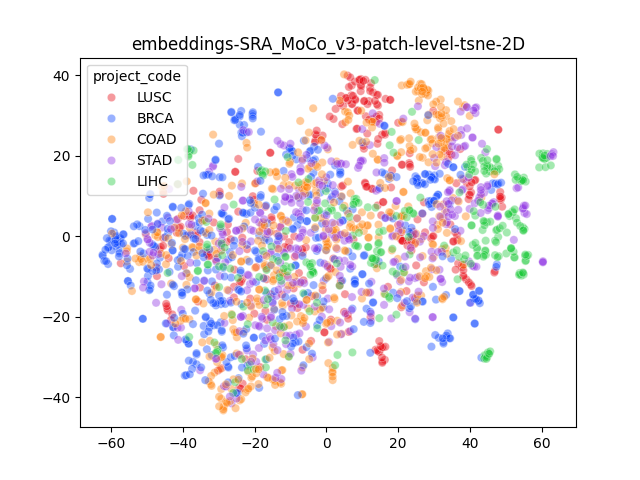} &
\includegraphics[width=0.34\linewidth]{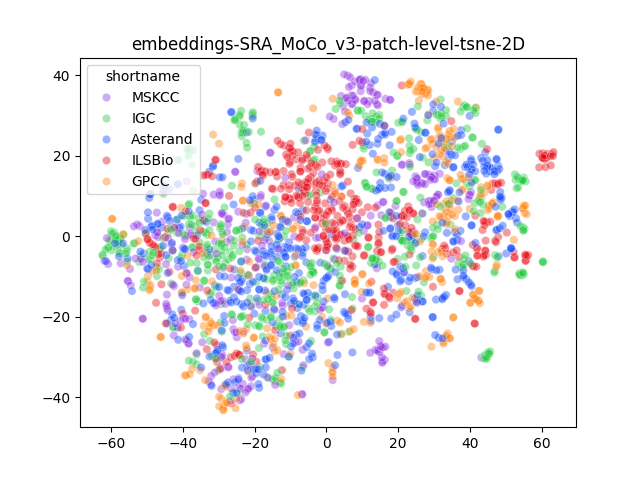} \\
\hline
\end{tabular}
\caption{Colorings of the t-SNE embeddings of all patches by cancer type (left) and medical center (right)}
\label{fig:colored_embeddings_1}
\end{table}
\renewcommand{\tablename}{Table}
\renewcommand{\thetable}{\arabic{table}}

\renewcommand{\tablename}{Figure}
\addtocounter{figure}{1}
\addtocounter{table}{-1}
\renewcommand{\thetable}{\arabic{figure}}
\begin{table}[H]
\centering
\begin{tabular}{|l|c|c|}
\hline
Virchow-1280D &
\includegraphics[width=0.34\linewidth]{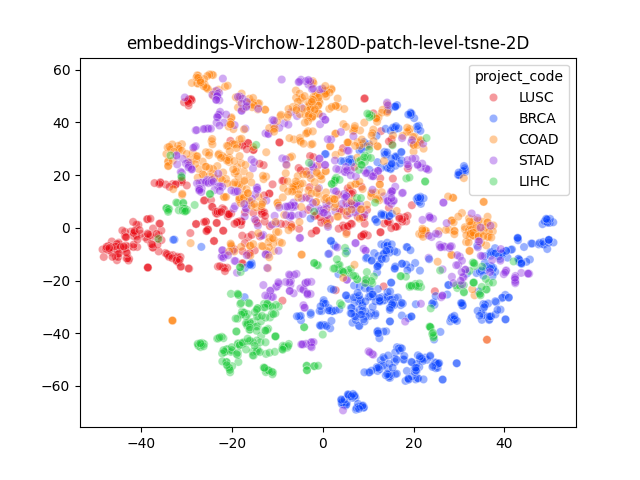} &
\includegraphics[width=0.34\linewidth]{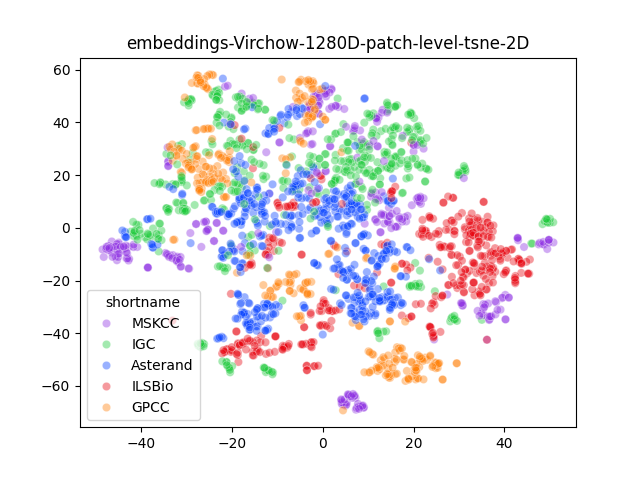} \\
\hline
Virchow-2560D &
\includegraphics[width=0.34\linewidth]{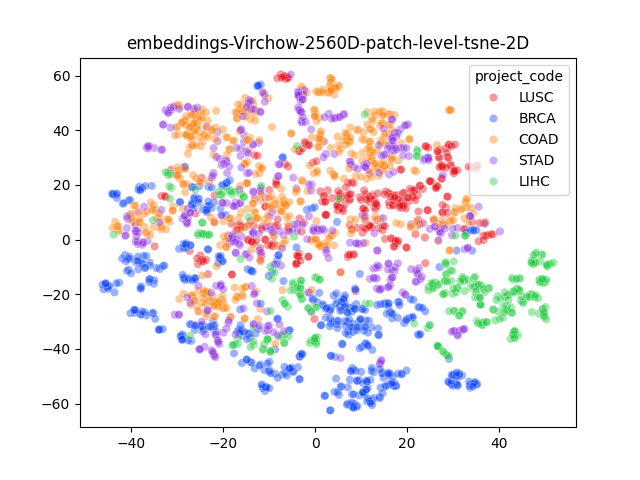} &
\includegraphics[width=0.34\linewidth]{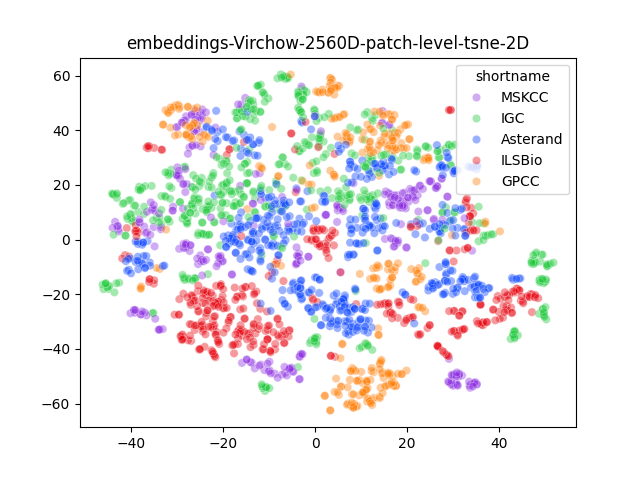} \\
\hline
Virchow2-1280D &
\includegraphics[width=0.34\linewidth]{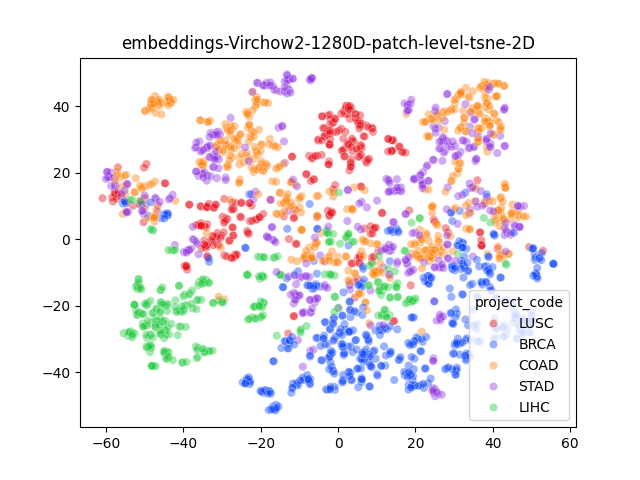} &
\includegraphics[width=0.34\linewidth]{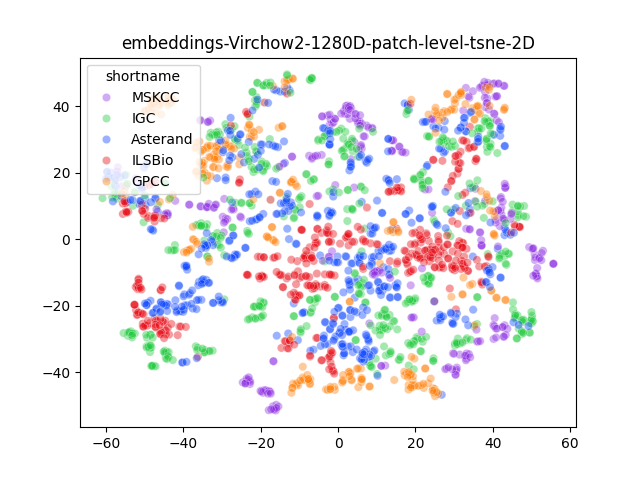} \\
\hline
Virchow2-2560D &
\includegraphics[width=0.34\linewidth]{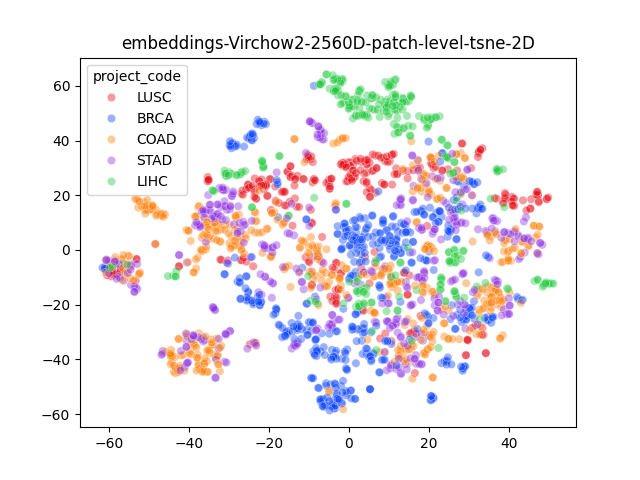} &
\includegraphics[width=0.34\linewidth]{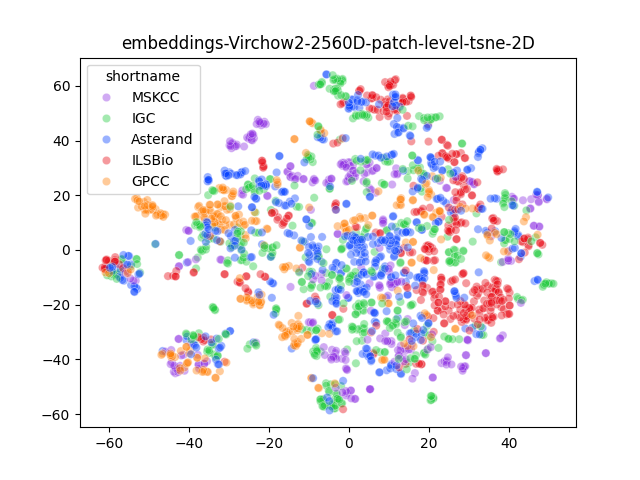} \\
\hline
H-optimus-0 &
\includegraphics[width=0.34\linewidth]{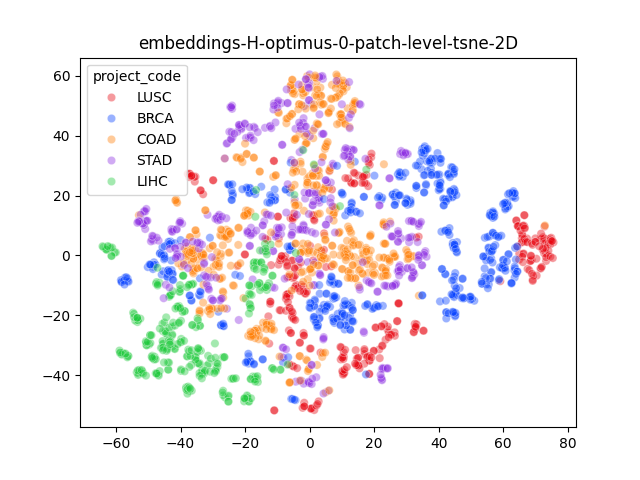} &
\includegraphics[width=0.34\linewidth]{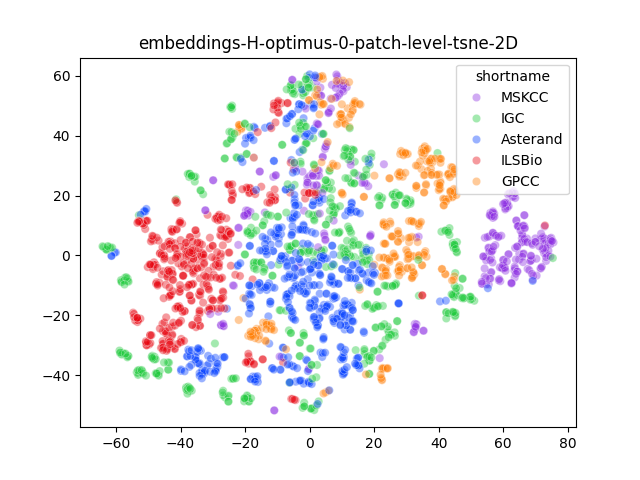} \\
\hline
prov-gigapath &
\includegraphics[width=0.34\linewidth]{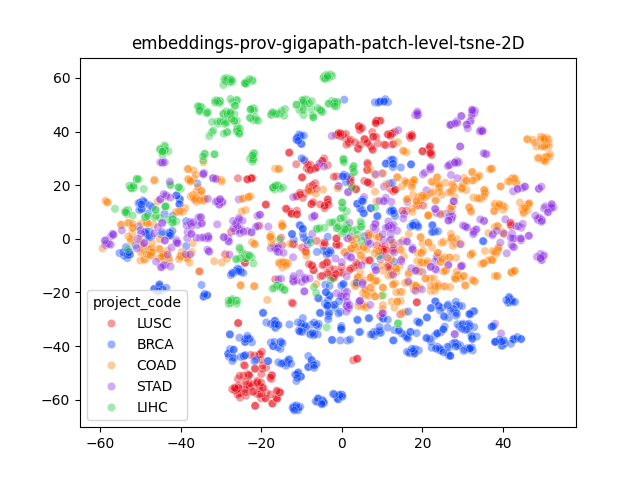} &
\includegraphics[width=0.34\linewidth]{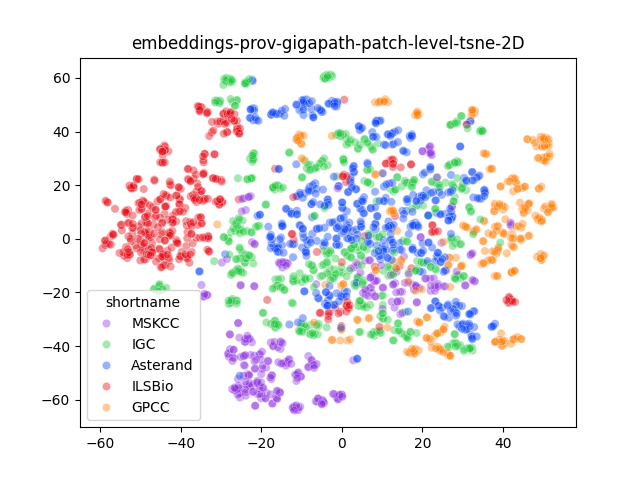} \\
\hline
\end{tabular}
\caption{Colorings of the t-SNE embeddings of all patches by cancer type (left) and medical center (right)}
\label{fig:colored_embeddings_2}
\end{table}
\renewcommand{\tablename}{Table}
\renewcommand{\thetable}{\arabic{table}}

\subsection{Relation between Prediction Performance and Robustness}
Ideally, a model should in our view demonstrate high prediction performance on relevant tasks, and at the same time show high robustness to irrelevant and confounding differences such as medical center differences. To evaluate what trade-off models achieve, we plot prediction performance on the cancer type classification task versus the prediction accuracy of the medical center, which relates inversely to robustness.

Figure \ref{fig:acc-ctype-center-full-embedding} shows the results for prediction of cancer type and medical center from embeddings. The top row shows prediction using \textit{knn} with \textit{k}=3 and logistic regression. For logistic regression (top right), we see that all models except SRA predict the medical center to a very high degree of accuracy: EXAONEPath, Phikon and Phikon-v2 have cross-validated accuracies of  0.987, 0.987 and 0.993 respectively, the latter approaching perfect center prediction. See Table \ref{tab:log-reg-accuracy-full-embeddings} for numerical results. 

The prediction performance for cancer type appears to be correlated with that for medical center; this raises the question whether high cancer type prediction accuracy is based on confounding medical center features. It is therefore questionable whether this prediction performance will generalize to unseen, new medical centers (Out Of Distribution evaluation).

For \textit{knn} on the full embeddings (top left), the accuracy of cancer type and medical center are reduced compared to logistic regression. There is a larger spread between the various results; and using \textit{knn}, there is one model, Virchow2, that performs better on cancer type prediction than on medical center prediction, indicating a better relation between biologically relevant prediction performance and robustness. The bottom two graphs show analogous results, but based on using 2D t-SNE coordinates as input rather than the full embeddings. 

\begin{table}[h]
\centering
\begin{tabular}{|l|c|c|c|c|}
\hline
\textbf{Model} & \textbf{Mean Accuracy} & \textbf{Std Dev} & \textbf{Mean Accuracy} & \textbf{Std Dev } \\
  & \textbf{Cancer Type} & \textbf{Cancer Type} & \textbf{Medical Center} & \textbf{Med.~Center} \\
\hline
SRA\_MoCo\_v3 & 0.486 & 0.036 & 0.692 & 0.027 \\
phikon & 0.829 & 0.037 & 0.987 & 0.012 \\
phikon-v2 & 0.83 & 0.038 & 0.993 & 0.007 \\
UNI & 0.713 & 0.034 & 0.956 & 0.013 \\
UNI2-h & 0.754 & 0.027 & 0.96 & 0.014 \\
hibou-b & 0.689 & 0.03 & 0.933 & 0.017 \\
Virchow-1280D & 0.727 & 0.038 & 0.932 & 0.022 \\
Virchow2-1280D & 0.786 & 0.03 & 0.957 & 0.016 \\
H-optimus-0 & 0.767 & 0.038 & 0.948 & 0.019 \\
prov-gigapath & 0.711 & 0.031 & 0.934 & 0.019 \\
EXAONEPath & 0.808 & 0.037 & 0.987 & 0.011 \\
\hline
\end{tabular}
\caption{Mean and standard deviation for the accuracy of cancer type prediction and medical center prediction using the full embedding vectors as input and logistic regression as the learning method.}
\label{tab:log-reg-accuracy-full-embeddings}
\end{table}

\renewcommand{\tablename}{Figure}
\addtocounter{figure}{1}
\addtocounter{table}{-1}
\renewcommand{\thetable}{\arabic{figure}}
\begin{table}[h]
\centering
\begin{tabular}{|c|c|}
\hline
    \includegraphics[width=0.5\linewidth]{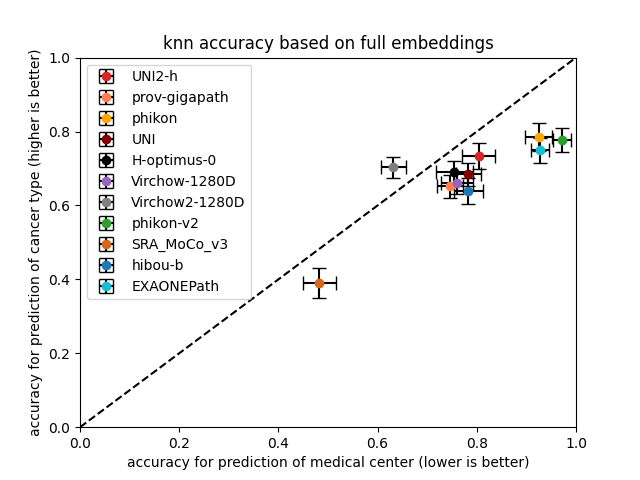}&
    \includegraphics[width=0.5\linewidth]{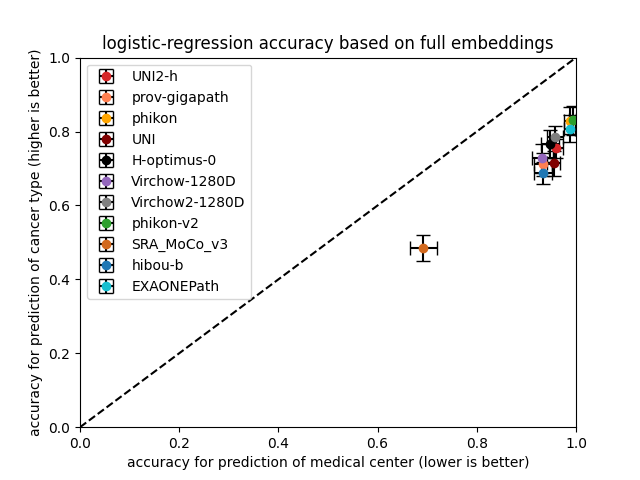}\\
    \includegraphics[width=0.5\linewidth]{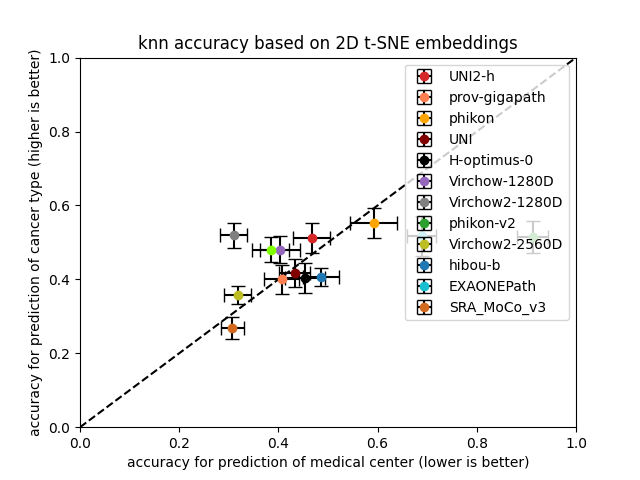}&
    \includegraphics[width=0.5\linewidth]{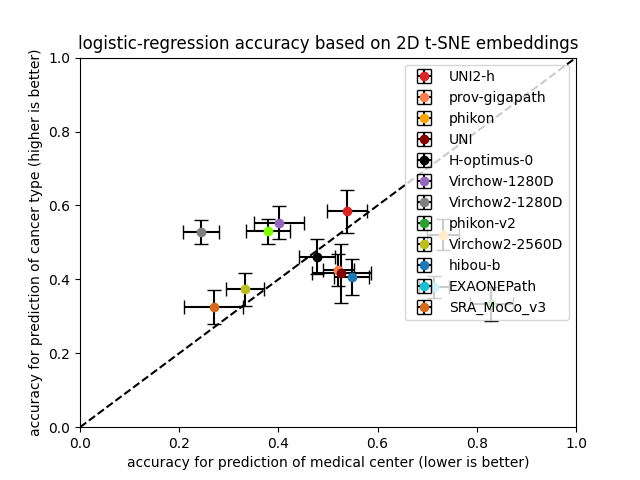}\\
\hline
\end{tabular}
\caption{
Top row: Accuracy of cancer type prediction vs.~center prediction when using the full embedding vectors as input using knn (left) and logistic regression (right).
Bottom row: Accuracy of cancer type prediction vs.~center prediction when using the 2D t-SNE embedding vectors as input using knn (left) and logistic regression (right).
}
\label{fig:acc-ctype-center-full-embedding}
\label{fig:acc-ctype-center-tsne-2D}
\end{table}
\renewcommand{\tablename}{Table}
\renewcommand{\thetable}{\arabic{table}}

\subsection{Effect of Medical Center Influences on  Regression}
It could be argued that the strong influence of medical centers on prediction performance observed above is restricted to downstream models that use all dimensions of the embedding space; and that models using regression can select those dimensions that code for biologically relevant features such as cancer type while ignoring dimensions encoding confounding information such as medical centers. 

\begin{figure}
    \centering
    \includegraphics[width=0.5\linewidth]{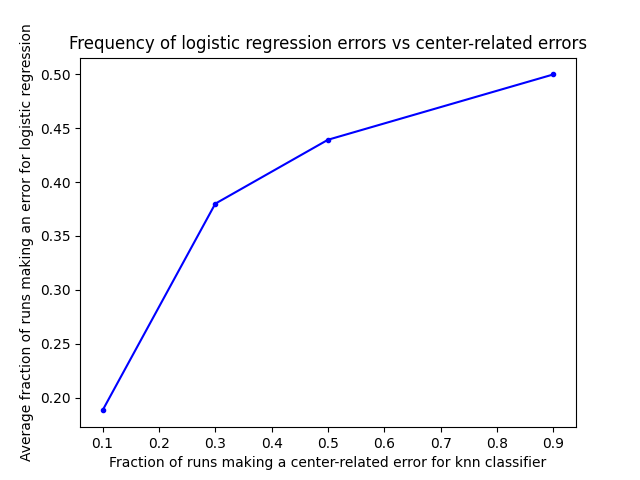}
    \caption{Relation between logistic regression errors and center-related knn errors. Samples that are more frequently misclassified by knn based on medical center are also more frequently misclassified by logistic regression, suggesting center similarities also affect logistic regression predictions.
     }
    \label{fig:log_reg}
\end{figure}

To test whether medical center influences affect logistic regression, the following analysis is performed. For each sample wrongly predicted by a logistic regression model, the fraction of knn runs making a center-related prediction error is calculated. A knn prediction error is considered to be center-related if the majority of its neighbors has:
\begin{itemize}
\item an incorrect class label prediction for the sample, and
\item the same medical center
\end{itemize}
Results are shown in Figure \ref{fig:log_reg}.

\section{Methods}
\subsection{Patch Extraction}

For patch extraction, WSITools \cite{wsitools} was used. We submitted PR \#11 to enable extracting a random selection of patches. Using this tool, patches of size 512x512 are extracted at the highest available resolution and downscaled by a factor 2 to 256x256. Given that the highest available resolution is typically 40X, this will typically result in a 20X resolution patch.

\begin{figure}[H]
    \centering
\includegraphics[width=0.5\linewidth]{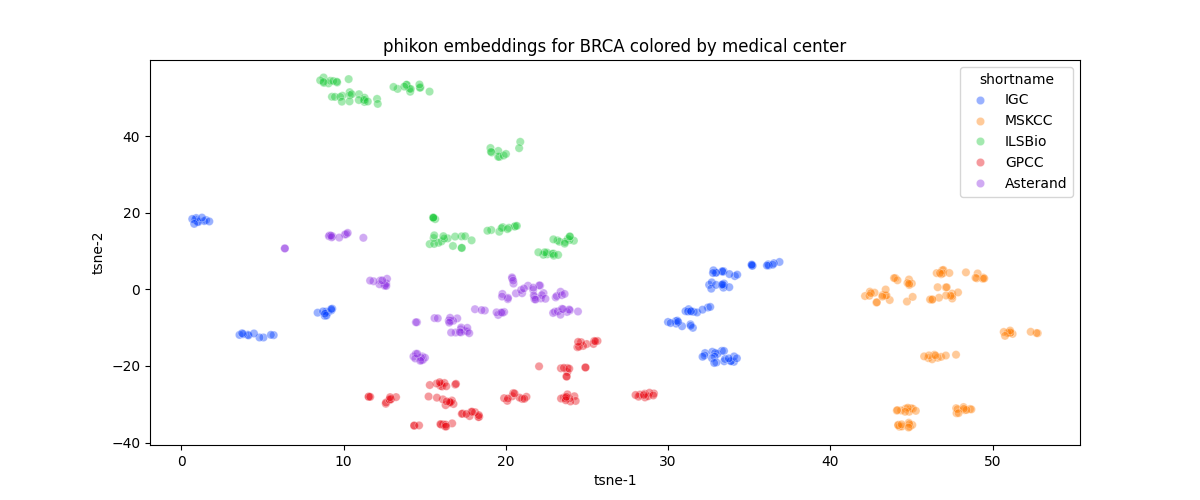}\hfill
\includegraphics[width=0.5\linewidth]{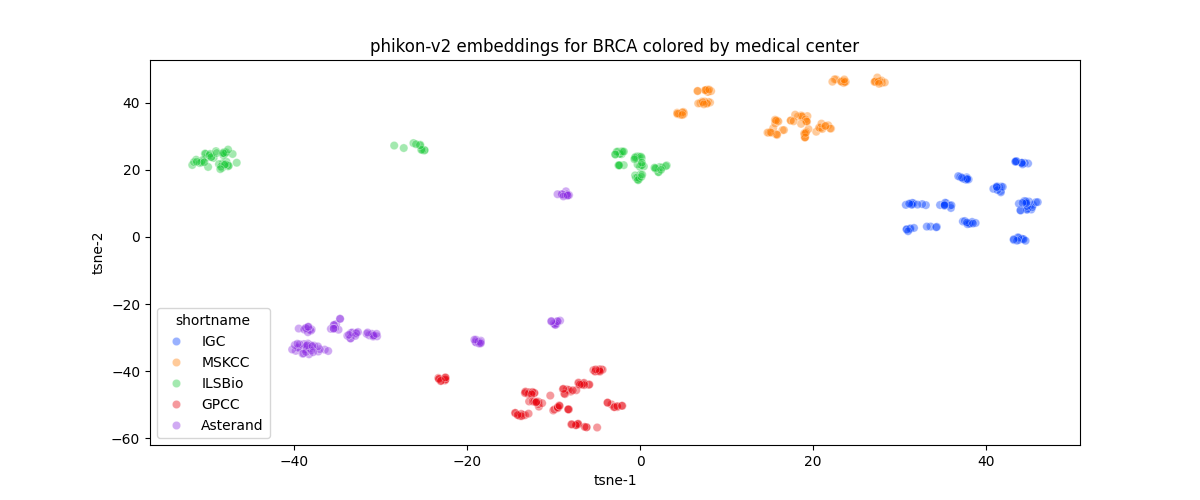}
    \caption{Embeddings for breast cancer colored by medical center for Phikon (left) and Phikon-v2 (right).}
    \label{fig:bc-center-phikon}
    \label{fig:bc-center-phikon-v2}
\end{figure}

\begin{figure}[H]
    \centering
\includegraphics[width=1.0\linewidth]{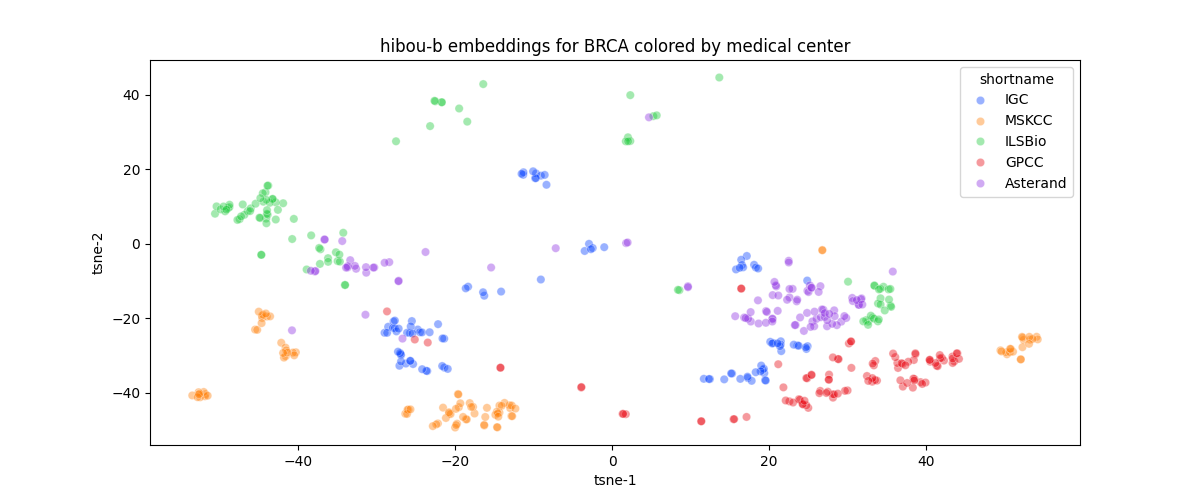}
    \caption{Embeddings for breast cancer colored by medical center for Hibou-B.}
    \label{fig:bc-center-hibou-b}
\end{figure}

\section{Discussion}
\subsection{Patch-level vs WSI-level Prediction}
Some patches may not contain sufficient information to determine the tissue of origin type / cancer type; thus, perfect classification may not be achievable at patch level, and higher levels of prediction accuracy may be achieved for an analogous WSI-level prediction task. The goal here however is not to maximize prediction accuracy, but rather to analyze the embedding space, and evaluate to what extent confounding center-related information influences classification decisions. A patch-level analysis provides the most direct way to link foundation model embeddings to medical centers; a WSI-level approach would introduce an extra level of indirection (e.g.~a MIL layer) between the foundation model and the downstream model output that would influence this relation and thus potentially obfuscate the analysis.

\subsection{Is Representation of Medical Center Information a Problem?}
It may be argued that SSL algorithms are designed to capture any differences between images, that differences between medical centers result in real differences between the images, and that it is therefore to be expected, or even desirable that pathology FMs learn to recognize, distinguish and represent medical centers. And one may attempt to reduce the influence of medical centers in post-hoc adaptations of the FM, or in the downstream model. 

Our belief however is that removing this influence is unlikely to be possible in an unbiased way; instead, it seems likely that the dimensions representing the medical center are not exactly orthogonal to dimensions representing biological information, and that it is therefore difficult or impossible to completely remove the influence of medical centers post hoc. In other words, medically relevant properties such as cancer risk are likely to be correlated with medical centers, as patient cohorts differ between medical centers.

Furthermore, it was seen that the prediction performance for cancer type appears to be correlated with that for medical center; this raises the question whether high cancer type prediction accuracy is based on confounding medical center features. It is therefore questionable whether this prediction performance will generalize to unseen, new medical centers (Out Of Distribution evaluation).

The application of AI in the medical domain brings with it a high degree of responsibility; if biases related to medical centers affect model predictions, and thereby influence patient diagnosis, treatment options, and outcomes, then it is the responsibility of practitioners in the medical AI domain to measure and reduce these influences to the maximal feasible extent.

\section{Conclusion}
In this work, robustness is viewed as insensitivity to confounding features. The Robustness Index, a novel metric to evaluate the degree to which biological information dominates confounding information such as the medical center, was introduced. Foundation models were seen to differ significantly in robustness according to this metric. Uni2-h and Virchow2 were found to be most robust, and Virchow2 was the only model so far with a robustness index above one, meaning biological information (cancer type) dominates confounding information (medical center) across the $k=50$ nearest neighbors.

It was seen that distance in embedding space strongly correlates with both the probability of encountering same-cancer-type neighbors and same-medical-center neighbors. This influence is not just local, but was seen to extend across the entire embedding space.

Using the notion of \textit{same-center confounders}, the impact of medical centers on prediction was evaluated, and it was found that all pathology foundation models evaluated here represent medical centers to a large extent.

A 2D projection of the embedding space was visualized. The resulting images show visually that the organization of the embedding space shows a clustering by medical center; more strongly so than a clustering by tissue or cancer type.

The robustness index and the other analysis techniques described in this work are intended as tools that may enable the development of more robust pathology foundation models.

\FloatBarrier 

\section{Appendix}
\label{appendix}
\subsection{Model Selection}
\label{app:model_selection}

Ten publicly available pathology foundation models were selected for evaluation, focusing on patch-level models. In addition, SRA\_MoCo\_v3 \cite{sra_moco_v3} was evaluated; while this model has been trained on a small single-tissue dataset, and can thus not be viewed as a foundation model, it is aimed at providing a more robust model. The selection consists of the following models:
\begin{itemize}
\item Phikon \cite{phikon}
\item Phikon-v2 \cite{phikonv2}
\item EXAONEPath \cite{exa_onepath} 
\item Prov\_gigapath \cite{prov_gigapath}
\item SRA\_MoCo\_v3 \cite{sra_moco_v3} 
\item UNI \cite{uni}
\item UNI2-h \cite{uni}
\item Hibou \cite{hibou}
\item H-Optimus-0 \cite{hoptimus0}
\item Virchow \cite{virchow}
\item Virchow2 \cite{virchow2}
\end{itemize}

\subsection{Embedding Generation}
Embeddings are generated using the default approach for each model. For Virchow and Virchow2, this means the average of the patch tokens is concatenated to the class token, resulting in a 2560-dimensional embedding, indicated with '-2560D'. To check whether this expansion of the embedding space affects performance, results with just the class token, indicated with '-1280D', are included for Virchow and Virchow2 as well. For all remaining models, the class token is the standard output, and is used here.\\


{\subsection{Fraction of Same-Center Confounders: Full Results}
\label{app:fraction-confounders-full}
\noindent
\vspace{-1cm}
\enlargethispage{4\baselineskip}
\renewcommand{\tablename}{Figure}
\addtocounter{figure}{1}
\addtocounter{table}{-1}
\renewcommand{\thetable}{\arabic{figure}}
\begin{table}[H]
\centering
\begin{tabular}{|c|c|}
\hline
\begin{minipage}[H]{0.36\textwidth}
    \centering
    \includegraphics[width=\linewidth]{fig/confounding-neighbors-fraction-same-center/confounding-neighbors-fraction-same-center-phikon-v2-nr_reps-5-nr_points-9.png}
\end{minipage}
&
\begin{minipage}[H]{0.36\textwidth}
    \centering
    \includegraphics[width=\linewidth]{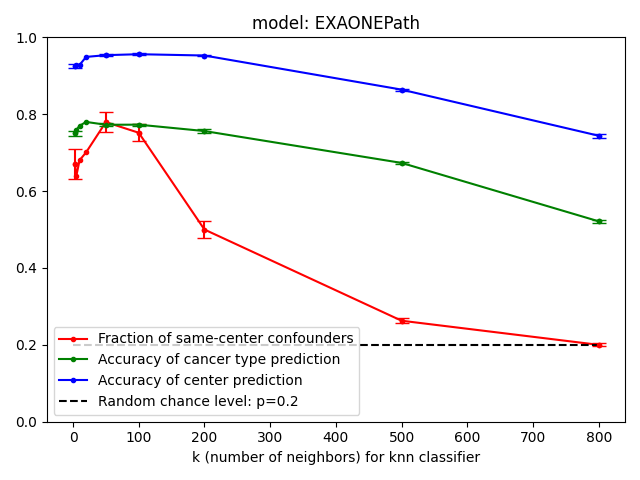}
\end{minipage}
\\[-8pt]
\hline
\begin{minipage}[H]{0.36\textwidth}
    \centering
    \includegraphics[width=\linewidth]{fig/confounding-neighbors-fraction-same-center/confounding-neighbors-fraction-same-center-phikon-nr_reps-5-nr_points-9.png}
\end{minipage}
&
\begin{minipage}[H]{0.36\textwidth}
    \centering
    \includegraphics[width=\linewidth]{fig/confounding-neighbors-fraction-same-center/confounding-neighbors-fraction-same-center-UNI-nr_reps-5-nr_points-9.png}
\end{minipage}
\\[-8pt]
\hline
\begin{minipage}[H]{0.36\textwidth}
    \centering
    \includegraphics[width=\linewidth]{fig/confounding-neighbors-fraction-same-center/confounding-neighbors-fraction-same-center-prov-gigapath-nr_reps-5-nr_points-9.png}
\end{minipage}
&
\begin{minipage}[H]{0.36\textwidth}
    \centering
    \includegraphics[width=\linewidth]{fig/confounding-neighbors-fraction-same-center/confounding-neighbors-fraction-same-center-H-optimus-0-nr_reps-5-nr_points-9.png}
\end{minipage}
\\[-8pt]
\hline
\begin{minipage}[H]{0.36\textwidth}
    \centering
    \includegraphics[width=\linewidth]{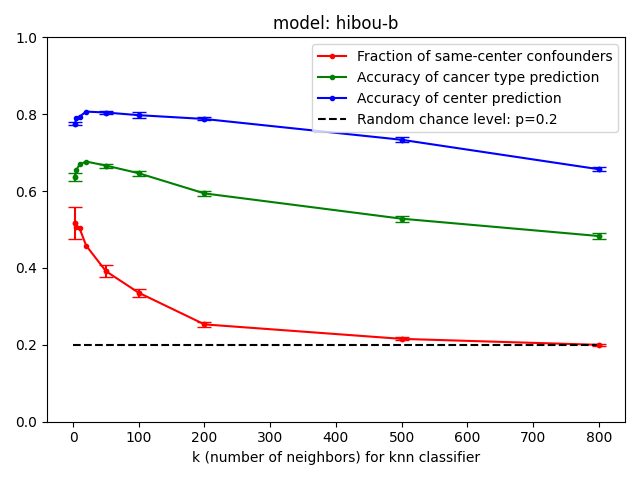}
\end{minipage}
&
\begin{minipage}[H]{0.36\textwidth}
    \centering
    \includegraphics[width=\linewidth]{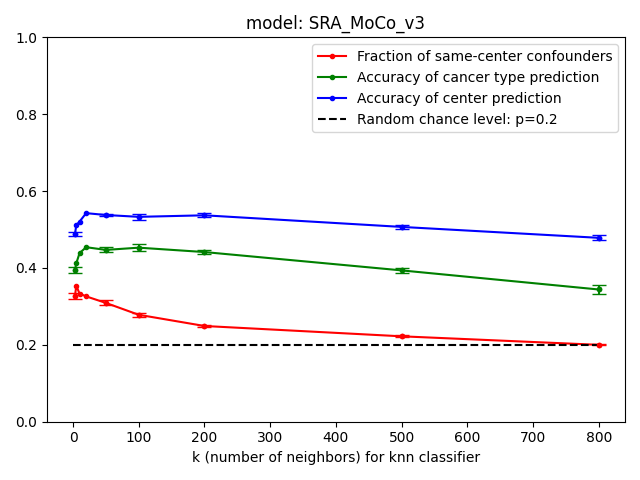}
\end{minipage}
\\[-8pt]
\hline
\begin{minipage}[H]{0.36\textwidth}
    \centering
    \includegraphics[width=\linewidth]{fig/confounding-neighbors-fraction-same-center/confounding-neighbors-fraction-same-center-Virchow-1280D-nr_reps-5-nr_points-9.png}
\end{minipage}
&
\begin{minipage}[H]{0.36\textwidth}
    \centering
    \includegraphics[width=\linewidth]{fig/confounding-neighbors-fraction-same-center/confounding-neighbors-fraction-same-center-Virchow2-1280D-nr_reps-5-nr_points-9.png}
\end{minipage}
\\[-8pt]
\hline
\begin{minipage}[H]{0.36\textwidth}
    \centering
    \includegraphics[width=\linewidth]{fig/confounding-neighbors-fraction-same-center/confounding-neighbors-fraction-same-center-UNI2-h-nr_reps-5-nr_points-9.png}
\end{minipage}
&
\begin{minipage}[H]{0.36\textwidth}
\end{minipage}\\
\hline
\end{tabular}
\caption{Fraction of same-center confounders: (i)  accuracy of tissue of origin / cancer type prediction (green), (ii) the accuracy of medical center prediction (blue), and (iii) fraction of  same-center confounders. All models show a substantial and significant influence of same-center confounders.}
\label{fig:fraction-confounders-full}
\end{table}
\renewcommand{\tablename}{Table}
\renewcommand{\thetable}{\arabic{table}}
}


{\subsection{Frequency Same Cancer Type / Medical Center: Full Results}
\label{app:freq-same-class-full}
\noindent
\vspace{-1cm}
\enlargethispage{4\baselineskip}
\renewcommand{\tablename}{Figure}
\addtocounter{figure}{1}
\addtocounter{table}{-1}
\renewcommand{\thetable}{\arabic{figure}}
\begin{table}[H]
\centering
\begin{tabular}{|c|c|}
\hline
\begin{minipage}[t]{0.36\textwidth}
    \centering
    \includegraphics[width=\linewidth]{fig/freq-same-class/freq-same-class-phikon-v2.png}
\end{minipage}
&
\begin{minipage}[t]{0.36\textwidth}
    \centering
    \includegraphics[width=\linewidth]{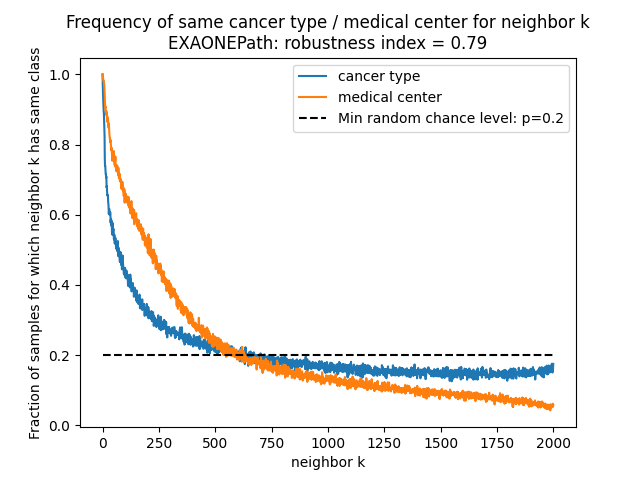}
\end{minipage}
\\[-8pt]
\hline
\begin{minipage}[t]{0.36\textwidth}
    \centering
    \includegraphics[width=\linewidth]{fig/freq-same-class/freq-same-class-phikon.png}
\end{minipage}
&
\begin{minipage}[t]{0.36\textwidth}
    \centering
    \includegraphics[width=\linewidth]{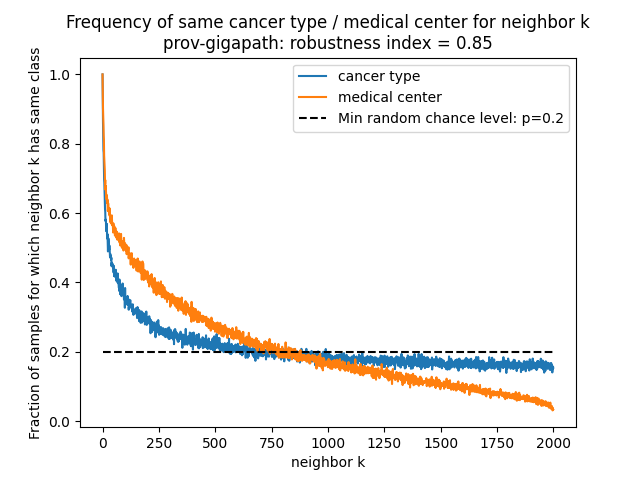}
\end{minipage}
\\[-8pt]
\hline
\begin{minipage}[t]{0.36\textwidth}
    \centering
    \includegraphics[width=\linewidth]{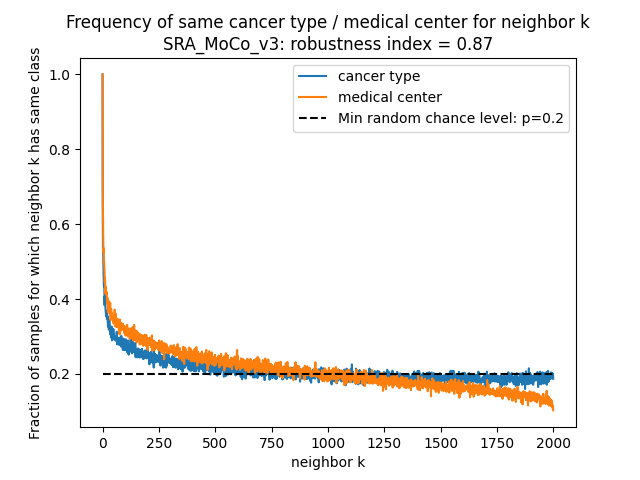}
\end{minipage}
&
\begin{minipage}[t]{0.36\textwidth}
    \centering
    \includegraphics[width=\linewidth]{fig/freq-same-class/freq-same-class-UNI.png}
\end{minipage}
\\[-8pt]
\hline
\begin{minipage}[t]{0.36\textwidth}
    \centering
    \includegraphics[width=\linewidth]{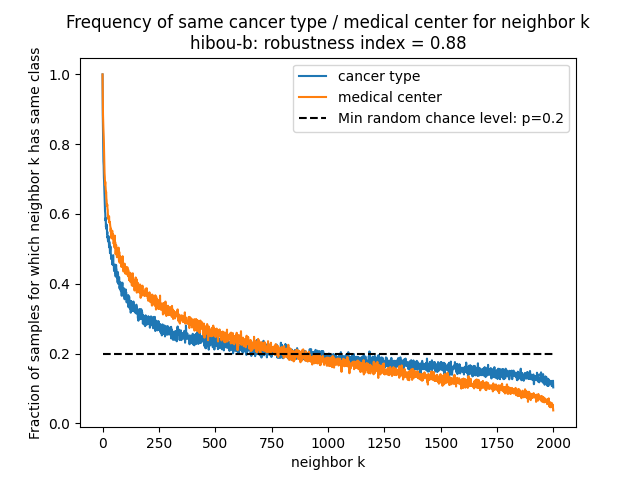}
\end{minipage}
&
\begin{minipage}[t]{0.36\textwidth}
    \centering
    \includegraphics[width=\linewidth]{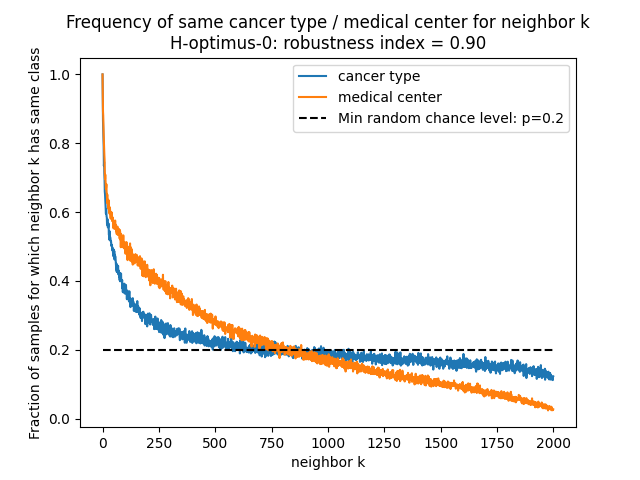}
\end{minipage}
\\[-8pt]
\hline
\begin{minipage}[t]{0.36\textwidth}
    \centering
    \includegraphics[width=\linewidth]{fig/freq-same-class/freq-same-class-Virchow-1280D.png}
\end{minipage}
&
\begin{minipage}[t]{0.36\textwidth}
    \centering
    \includegraphics[width=\linewidth]{fig/freq-same-class/freq-same-class-UNI2-h.png}
\end{minipage}
\\[-8pt]
\hline
\begin{minipage}[t]{0.36\textwidth}
    \centering
    \includegraphics[width=\linewidth]{fig/freq-same-class/freq-same-class-Virchow2-1280D.png}
\end{minipage}
&
\begin{minipage}[t]{0.36\textwidth}
\end{minipage}
\\[-8pt]
\hline
\end{tabular}
\caption{Fraction of samples for which the k-th neighbor has the same cancer type (blue) or medical center (orange), in order of increasing robustness}
\label{app:fraction-same-class}
\end{table}
\renewcommand{\tablename}{Table}
\renewcommand{\thetable}{\arabic{table}}
\nopagebreak[4]}

\section{Online resources}
We intend to make the patch dataset constructed and used in this work available online. An extended version of this work, combined with related simultaneous research from the TU Berlin BIFOLD group and Aignostics, is in preparation. 

\section{Acknowledgements}
The authors would like to thank Hans Pinckaers, Jonas Dippel, Alexander Möllers, Maximilian Alber, other colleagues at Aignostics, and the TU Berlin BIFOLD group for valuable suggestions that improved the article.
The results presented here are based upon data generated by the TCGA Research Network: \href{http://cancergenome.nih.gov}{http://cancergenome.nih.gov}. We thank Bodong Zhang for kindly providing the SRA\_MoCo\_v3 model for inclusion in this evaluation.
\FloatBarrier 

\bibliographystyle{unsrt}
\bibliography{bibfile}

\end{document}